\documentclass[runningheads]{llncs}

\usepackage{eccv}
\usepackage{eccvabbrv}

\usepackage{graphicx}
\usepackage{booktabs}

\usepackage[accsupp]{axessibility}

\usepackage{bm}
\usepackage{subcaption}
\usepackage{ifthen}
\usepackage{pifont}
\usepackage{siunitx}

\usepackage{float}
\usepackage{tabularx}
\newcolumntype{Y}{>{\centering\arraybackslash}X}
\usepackage{multirow}

\usepackage{tikz}
\usetikzlibrary{backgrounds,arrows.meta,positioning}

\usepackage{pgfplots}
\pgfplotsset{compat=1.18}

\usepackage{xcolor}
\definecolor{RWTHBlau}{RGB}{0,84,159}
\definecolor{RWTHMagenta}{RGB}{227,0,102}
\definecolor{RWTHGelb}{RGB}{255,237,0}
\definecolor{RWTHPetrol}{RGB}{0,97,101}
\definecolor{RWTHTuerkis}{RGB}{0,152,161}
\definecolor{RWTHGruen}{RGB}{87,171,39}
\definecolor{RWTHMaigruen}{RGB}{189,205,0}
\definecolor{RWTHOrange}{RGB}{246,168,0}
\definecolor{RWTHRot}{RGB}{204,7,30}
\definecolor{RWTHBordeaux}{RGB}{161,16,53}
\definecolor{RWTHViolett}{RGB}{97,33,88}
\definecolor{RWTHLila}{RGB}{122,111,172}

\newcommand{\up}{$\uparrow$}
\newcommand{\down}{$\downarrow$}

\newcolumntype{s}{>{\centering}p}

\newcommand{\PAR}[1]{\vskip4pt \noindent {\bf #1~}}
\newcommand{\PARbegin}[1]{\noindent {\bf #1~}}

\newcommand{\best}[1]{\textbf{#1}}
\newcommand{\secn}[1]{\underline{#1}}
\newcommand{\hide}[1]{\textcolor{black!45}{#1}}
\newcommand{\bestnaive}[1]{\colorbox{RWTHGelb!30}{#1}}
\newcommand{\bestpolicy}[1]{\colorbox{RWTHGruen!30}{#1}}

\makeatletter
\newcommand\smalltext{\@setfontsize\smalltext{8}{9}}
\makeatother

\DeclareMathOperator*{\argmax}{arg\,max}

\pgfplotsset{
  tight and small labels/.style={
    xlabel style={
      font=\scriptsize,
      inner sep=0pt,
      outer sep=1pt,
    },
    ylabel style={
      font=\scriptsize,
      inner sep=0pt,
      outer sep=1pt,
    },
    ticklabel style={
      font=\scriptsize,
      inner sep=0pt,
      outer sep=1pt,
    },
  },
}
\tikzset{
    trajectory/.style={-{Stealth}},
}

\newcommand{\httpsurl}[1]{\href{https://#1}{\texttt{#1}}}

\newcommand{\policyarrow}[3][0pt]{%
  \makebox[#1][r]{#2} $\rightarrow$ \makebox[#1][l]{#3}%
}

\usepackage{hyperref}

\usepackage{orcidlink}

\begin{document}

\title{Towards Metric-Agnostic Trajectory Forecasting}

\author{Markus Knoche\inst{1}\orcidlink{0009-0000-1256-0978} \and
Daan de Geus\inst{2}\orcidlink{0000-0003-0559-5341} \and
Bastian Leibe\inst{1}\orcidlink{0000-0003-4225-0051}}

\authorrunning{M.~Knoche et al.}

\institute{RWTH Aachen University, Germany\\
\and Eindhoven University of Technology, Netherlands\\
\httpsurl{vision.rwth-aachen.de/TraDiE-policies}}

\maketitle

\begin{abstract}
    Accurate trajectory forecasting of surrounding traffic participants is a core capability for autonomous driving, enabling vehicles to anticipate behavior and plan safe maneuvers. We observe that current state-of-the-art forecasting models on Argoverse~2 and the Waymo Open Motion Dataset tailor their training objectives to the different benchmark metrics. Because these metrics encourage conflicting behavior, we propose a paradigm change for trajectory forecasting: \emph{training models with metric-agnostic probabilistic objectives and treating metric optimization as a downstream task applied to the predictive distribution}. Concretely, we introduce Trajectory Distribution Evaluation (TraDiE) policies, metric-specific policies that map a predictive distribution to the set of $K$ trajectories and confidences required by trajectory forecasting metrics. We evaluate this framework by introducing DONUT-NLL, which adapts the training objective of the state-of-the-art trajectory forecasting model DONUT to directly optimize the predictive distribution. Using our policies, DONUT-NLL achieves state-of-the-art results on all metrics of the Waymo motion prediction benchmark.
\end{abstract}

\section{Introduction}

Trajectory forecasting is a core component of autonomous driving systems, as it enables vehicles to anticipate the behavior of other traffic participants. Because human driving is inherently non-deterministic and multimodal, it is not sufficient to predict only a single future trajectory per agent. Instead, trajectory forecasting models must capture multiple plausible futures and quantify uncertainty such that downstream applications can account for possible behaviors.

In practice, the trajectory forecasting task is defined as follows: given a road graph describing the static scene context (\eg, lanes, crosswalks) and the recent state histories of relevant agents (\eg, cars, pedestrians), the model must predict a set of $K$ future trajectories over a fixed temporal horizon for each agent. Typically, $K=6$. In addition, each trajectory gets assigned a confidence score. Together, these predictions should represent a multimodal distribution over plausible futures corresponding to the true uncertainty of the world.

In recent years, a diverse set of architectures and training objectives has been proposed for this task \cite{knoche2025donut, zhou2023qcnet, shi2024mtrpp, sun2025impact, huang2025drivegpt, zhou2025modeseq}. Interestingly, we observe that these models are strongly shaped by the particular metrics that are used in the most popular benchmarks. Models evaluating on Argoverse~2 \cite{wilson2021argoverse2}, whose main metric \emph{Brier-minFDE} depends on endpoint distance, often employ training objectives that closely mirror these distance-based metrics \cite{zhou2022hivt, lan2024sept}. Conversely, methods evaluated on the Waymo benchmark \cite{ettinger2021waymo}, which uses \emph{soft mAP}, are optimized for broader coverage of the space of possible future endpoints, \eg, by using non-maximum suppression to avoid clusters of endpoints \cite{shi2022mtr, lin2024eda, yan2025trajflow}. Some recent papers even train multiple models to optimize different metrics \cite{shi2022mtr, sun2025rmpyolo, sun2025impact}.

While it is not inherently problematic to optimize for a benchmark, tightly coupling trajectory forecasting models to evaluation metrics can lead to several issues. Different benchmark metrics encode different, sometimes conflicting, goals: distance-based metrics reward precise final positions, whereas soft mAP expects broad coverage of the future space. As a result, models tuned to one metric tend to perform worse on others, and it becomes hard to assess whether improvements stem from genuinely better forecasting or from better metric-specific tuning. Moreover, benchmark metrics do not directly evaluate the model's predictive distribution and thus only provide a partial view of its quality.

In this work, we argue for a paradigm change in trajectory forecasting: \emph{the primary training objective should be to learn a good predictive distribution, independent of the metrics used to evaluate the forecast.} Such a distribution should provide a faithful representation of uncertainty. If it is well-calibrated, it can serve as a general, metric-agnostic interface that can then be exploited by various downstream tasks, including optimizing for evaluation metrics.

We suggest that optimizing for different metrics should become part of the evaluation protocol as a post-processing step applied to the generated predictive distribution. To this end, we build upon and generalize the per-metric optimization proposed by HOME \cite{gilles2021home} by defining metric-specific policies that map these distributions to the finite set of $K$ trajectories and confidences required by a particular metric. Our \textbf{Tra}jectory-\textbf{Di}stribution \textbf{E}valuation policies (TraDiE policies) are derived from Monte Carlo samples of the predictive distribution for common trajectory prediction metrics. Crucially, these policies are applied to the outputs of a single underlying model, without retraining. This evaluation protocol rewards the model for a good predictive distribution: if it is accurate, policies optimizing for the metric under the distribution will yield good performance.

We argue that adopting this decoupled view will allow the trajectory forecasting community to again focus on the core of the problem, namely how to represent the predictive distribution and how to improve its accuracy. This turns the choice of predictive distribution into a general design decision. It allows systematic evaluation of how future agent trajectories and their uncertainty should be represented and how trajectory forecasting models should be trained, independent of specific benchmark metrics.

As a concrete instantiation, we apply this idea to DONUT \cite{knoche2025donut}, a recent trajectory forecasting model that achieves state-of-the-art results on Argoverse~2. Naively applied to Waymo with its original, minFDE-related loss, DONUT's soft mAP performance is extremely poor, making it an ideal testbed to study the impact of metric-agnostic training.
We replace the original loss with negative log-likelihood objectives, which directly optimize the predictive distribution, and find that these models improve both minFDE and (soft) mAP significantly when combined with our policies. This shows that optimizing the distribution directly is indeed better than training for surrogate submetrics.
We further evaluate different distribution families for representing agents' positions and show that mixtures of generalized Gaussian distributions are better suited than mixtures of Laplacian distributions or Gaussian scale mixtures.
Beyond DONUT, we apply our metric-specific policies to two strong open-source baselines (MTR~\cite{shi2022mtr} and QCNet~\cite{zhou2023qcnet}) and show that they (i) consistently improve MTR across metrics and (ii) expose a distribution mismatch for QCNet under distance-based evaluation.

In summary, our contributions are:
\begin{enumerate}
    \item We advocate for a paradigm shift in which trajectory forecasting models are trained with metric-agnostic probabilistic objectives that learn well-calibrated predictive distributions, and in which metric optimization is treated as a downstream step via metric-specific policies.
    \item We propose sampling-based metric-specific policies for mapping predictive distributions to $K$ trajectories and confidences, with concrete implementations for minFDE, (soft) mAP and miss rate, enabling evaluation of multiple metrics from a single predictive model without metric-specific retraining.
    \item We introduce DONUT-NLL, a negative log-likelihood training scheme for DONUT, which achieves state-of-the-art results on the Waymo benchmark, demonstrating the effectiveness of optimizing the distribution directly.
    \item We conduct a systematic comparison of position distributions and empirically find that generalized Gaussian distributions outperform alternatives.
    \item We validate that the proposed policies are model-agnostic by applying them to MTR~\cite{shi2022mtr} and QCNet~\cite{zhou2023qcnet}, highlighting both gains and failure modes when predictive distributions are misspecified.
\end{enumerate}

\section{Related Work}

Early work on trajectory forecasting encodes the environment in bird's-eye-view rasterizations and applies convolutional neural networks to capture local context \cite{lee2017desire, gilles2021home, chai2020multipath, cui2019multimodal, phan2020covernet, salzmann2020trajectron++, hong2019rules}. More modern approaches use sparse input representations, where agents' histories and map elements are modeled as polylines \cite{liang2020lanegcn, gao2020vectornet}. On top of such structures inputs, graph neural networks have been used to capture relations between agents and the road layout \cite{mohamed2020social-stgcnn, zeng2021lanercnn}. Current approaches commonly model agents as queries and use attention mechanisms for interaction between agents and map elements \cite{liu2021mmtransformer, lin2024eda, nayakanti2023wayformer, shi2022mtr, sun2025rmpyolo, song2025unimotion, ruan2024learning, zhou2023qcnet, knoche2025donut, shi2024mtrpp, wang2025futurenetlof, zhou2025modeseq}.

To model uncertainty over future trajectories, earlier works \cite{hong2019rules, lee2017desire} have employed VAEs \cite{kingma2014vae}, which allow sampling of an arbitrary number of trajectories. Some approaches \cite{gilles2021home, gilles2022gohome, gilles2022thomas} represent the uncertainty as dense, rasterized heatmaps. Most current models predict a fixed number of possible futures, often represented as the modes of Gaussian mixtures \cite{shi2022mtr, shi2024mtrpp, shi2024mtrv3, sun2025rmpyolo, nayakanti2023wayformer, shi2022mtra, sun2025impact, lin2024eda, yan2025trajflow} or Laplacian mixtures \cite{zhou2022hivt, knoche2025donut, wang2025futurenetlof, zhou2023qcnet, zhou2025modeseq}. For training, they commonly employ a winner-takes-all loss, in which the trajectory closest to the ground truth gets optimized with negative log-likelihood. Other approaches \cite{zhang2024demo, lan2024sept, zhang2025polaris} directly regress the best trajectory with an $L_1$ or smooth $L_1$ loss without defining a continuous mixture. A few works \cite{seff2023motionlm, huang2025drivegpt} employ GPT-like models, which discretize sub-trajectories into a fixed number of classes and learn a categorical distribution per timestep. Such models allow sampling arbitrarily many trajectories at inference time.

As we discuss in more detail in \cref{sec:metric_train}, many of these models optimize for the main metrics of either Argoverse~2 or Waymo. As a result, models perform well on specific aspects of the trajectory prediction task, rather than improving the predictive distribution itself. Moreover, the objectives induced by these metrics are partly conflicting, so models optimized for one benchmark metric typically perform poorly on the other. In contrast, we propose to train models with metric-agnostic probabilistic objectives that focus on learning good predictive distributions, and to handle different benchmark metrics via metric-specific policies applied at evaluation time, allowing a single calibrated model to be reused across multiple metrics and downstream tasks.

The idea to decouple predictive distribution estimation from reporting has also been explored by HOME \cite{gilles2021home}. HOME outputs a dense heatmap for the endpoints and introduces a mechanism to post-process this output separately for individual metrics. However, the post-processing is tightly coupled to a specific discrete endpoint representation and cannot be used as a general, model-agnostic solution. In contrast, we derive sampling-based policies that can be applied to any probabilistic forecaster and directly follow the metric definitions without model-specific tuning. We provide a detailed comparison to HOME in App.~\ref{app:home}.

\section{Evaluation Metrics}

In this section, we will present common evaluation metrics for trajectory forecasting and analyze how previous methods have implicitly or explicitly adapted their optimization pipeline for either distance-based or window-based metrics. For each agent $n$, these metrics evaluate a set of $K$ predicted trajectories $\{\hat{\bm X}_{nk}^\text{pos}\}_{k=1}^K$ where each trajectory $\hat{\bm X}_{nk}^\text{pos}$ is a sequence $(\hat{\bm x}_{nkt}^\text{pos})_{t=1}^T$ of 2D positions over $T$ future timesteps, given a ground-truth trajectory $\bm Y_{n}^\text{pos} = (\bm y_{nt}^\text{pos})_{t=1}^T$. Some metrics additionally consider predicted confidences $\{\hat\pi_{nk}\}_{k=1}^K$ assigned to the trajectories.

\subsection{Distance-Based Metrics}

\begin{figure}[t]
    \def\panelw{2.5}
    \def\panelh{2.0}
    \def\xsep{0.3}
    \def\ysep{0.3}
    \def\windoww{0.8}
    \def\windowh{1.6}
    \def\gtx{1.2}
    \def\gty{1.0}
    \def\gth{-20}
    \centering
    \begin{subfigure}{0.24\textwidth}
        \centering
        \begin{tikzpicture}
            \draw[black] (0,0) rectangle (\panelw, \panelh);
            
            \draw[trajectory, thick, black!30] (1.1, 0) to[out=90, in=240] node[anchor=west, black, pos=1] {\tiny $0.5$} (1.7, 1.5);
            \draw[trajectory, thick, black!30] (1.5, 0) to[out=90, in=240] node[anchor=west, black, pos=1] {\tiny $0.3$} (1.9, 1.0);
            \draw[trajectory, thick, black!30] (0.7, 0) to[out=90, in=260] node[anchor=east, black, pos=1, xshift=-1pt] {\tiny $0.3$} (0.9, 1.05);
            \draw[trajectory, ultra thick, RWTHGruen] (0.9, 0) to[out=90, in=250] (\gtx, \gty);
        \end{tikzpicture}
        \caption{predictions}
    \end{subfigure}
    \begin{subfigure}{0.24\textwidth}
        \centering
        \begin{tikzpicture}
            \draw[black] (0,0) rectangle (\panelw, \panelh);
            
            \draw[trajectory, thick, black!30] (1.1, 0) to[out=90, in=240] (1.7, 1.5);
            \draw[trajectory, thick, black!30] (1.5, 0) to[out=90, in=240] (1.9, 1.0);
            \draw[trajectory, thick, RWTHBlau] (0.7, 0) to[out=90, in=260] (0.9, 1.05);
            \draw[trajectory, ultra thick, RWTHGruen] (0.9, 0) to[out=90, in=250] (\gtx, \gty);
            
            \draw[thick] (1.2, 1.0) -- (1.216, 1.1);
            \draw[thick] (1.208, 1.05) -- (0.908, 1.1);
            \draw[thick] (0.9, 1.05) -- (0.916, 1.15);
        \end{tikzpicture}
        \caption{minFDE}
        \label{fig:minfde}
    \end{subfigure}
    \begin{subfigure}{0.24\textwidth}
        \centering
        \begin{tikzpicture}
        \draw[black] (0,0) rectangle (\panelw, \panelh);
        
        \draw[trajectory, thick, RWTHBlau] (1.1, 0) to[out=90, in=240] node[anchor=east, black, pos=1, xshift=2pt, yshift=1pt] {\tiny $+$} (1.7, 1.5);
        \draw[trajectory, thick, RWTHRot] (1.5, 0) to[out=90, in=240] node[anchor=west, black, pos=1, xshift=-2pt, yshift=1pt] {\tiny $-$} (1.9, 1.0);
        \draw[trajectory, thick, RWTHOrange] (0.7, 0) to[out=90, in=260] node[anchor=west, black, pos=1, xshift=-3pt, yshift=1pt] {$\circ$} (0.9, 1.05);
        \draw[trajectory, ultra thick, RWTHGruen] (0.9, 0) to[out=90, in=250] (\gtx, \gty);
        
        \draw[black, rotate around={\gth:(\gtx, \gty)}]
            (\gtx-.5*\windoww,\gty-.5*\windowh) rectangle
            (\gtx+.5*\windoww,\gty+.5*\windowh);
        \end{tikzpicture}
        \caption{(soft) mAP}
        \label{fig:map}
    \end{subfigure}
    \begin{subfigure}{0.24\textwidth}
        \centering
        \begin{tikzpicture}
            \begin{axis}[
                height=3.35cm,
                width=3.85cm,
                ymin=0, ymax=1.1,
                xmin=0, xmax=1.1,
                axis lines=left,
                xlabel={recall},
                ylabel={precision},
                xtick={0,1},
                ytick={0,1},
                tight and small labels,
                xlabel style={yshift=0.2cm,xshift=-0.1cm},
                ylabel style={yshift=-0.2cm,xshift=-0.1cm}
            ]

            \addplot [
                thick,
                fill=RWTHGruen!10,
            ] coordinates {
                (0.00, 1.00)
                (0.10, 1.00)
                (0.10, 0.95)
                (0.33, 0.95)
                (0.33, 0.80)
                (0.46, 0.80)
                (0.46, 0.75)
                (0.59, 0.75)
                (0.59, 0.72)
                (0.69, 0.72)
                (0.69, 0.55)
                (0.86, 0.55)
                (0.86, 0.42)
                (0.91, 0.42)
                (0.91, 0.22)
                (0.91, 0.22)
                (0.91, 0.08)
                (1.00, 0.08)
                (1.00, 0.00)
            } \closedcycle;

            \addplot [
                thick,
                RWTHBlau,
            ] coordinates {
                (0.00, 1.00)
                (0.10, 1.00)
                (0.10, 0.80)
                (0.15, 0.88)
                (0.22, 0.92)
                (0.33, 0.95)
                (0.33, 0.75)
                (0.46, 0.80)
                (0.46, 0.63)
                (0.54, 0.72)
                (0.59, 0.75)
                (0.59, 0.63)
                (0.64, 0.68)
                (0.69, 0.72)
                (0.69, 0.45)
                (0.75, 0.49)
                (0.83, 0.53)
                (0.86, 0.55)
                (0.86, 0.37)
                (0.91, 0.42)
                (0.91, 0.19)
                (0.91, 0.22)
                (0.91, 0.05)
                (1.00, 0.08)
                (1.00, 0.00)
            };

            \addplot [
                thick,
                RWTHGruen,
            ] coordinates {
                (0.00, 1.00)
                (0.10, 1.00)
                (0.10, 0.95)
                (0.33, 0.95)
                (0.33, 0.80)
                (0.46, 0.80)
                (0.46, 0.75)
                (0.59, 0.75)
                (0.59, 0.72)
                (0.69, 0.72)
                (0.69, 0.55)
                (0.86, 0.55)
                (0.86, 0.42)
                (0.91, 0.42)
                (0.91, 0.22)
                (0.91, 0.22)
                (0.91, 0.08)
                (1.00, 0.08)
                (1.00, 0.00)
            };
            
            \end{axis}
        \end{tikzpicture}
        \caption{smoothing}
        \label{fig:smooth}
    \end{subfigure}
  \caption{\textbf{Illustration of metrics.} \textbf{(a)} Each \textbf{\color{black!50} prediction} has a confidence assigned. \textbf{(b)} For the distance-based metric minFDE, the endpoint distance between \textbf{\color{RWTHGruen} ground truth} and \textbf{\color{RWTHBlau} closest prediction} is calculated. \textbf{(c)} For the window-based metric (soft) mAP, an oriented window is placed around the ground-truth endpoint. The \textbf{\color{RWTHBlau} highest-confidence trajectory within the window} ($+$) counts as a true positive, \textbf{\color{RWTHOrange} additional trajectories within the window} ($\circ$) are either counted as false positives (mAP) or ignored (soft mAP). \textbf{\color{RWTHRot} Predictions outside the window} ($-$) count as false positives. \textbf{(d)} A \textbf{\color{RWTHBlau} precision-recall curve} is created for each confidence score over the entire dataset and then \textbf{\color{RWTHGruen} smoothed}; the area under the curve is the (soft) mAP.}
\end{figure}
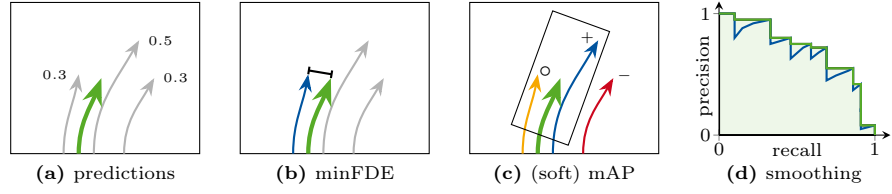

Distance-based metrics evaluate how close a predicted trajectory comes to the ground-truth trajectory. A common example is the minimum final displacement error (minFDE), visualized in \cref{fig:minfde}. For an agent $n$, it is defined as the Euclidean distance between the ground-truth endpoint $\bm y_{nT}^\text{pos}$ and the closest endpoint of $K$ predictions $\{\hat{\bm x}_{nkT}^\text{pos}\}_{k=1}^K$:
\begin{equation}
    \text{minFDE}(\bm y_{nT}^\text{pos}, \{\hat{\bm x}_{nkT}^\text{pos}\}_{k=1}^K) = \min_k \|\bm y_{nT}^\text{pos} - \hat{\bm x}_{nkT}^\text{pos}\|_2.
    \label{eq:minfde}
\end{equation}
The minFDE for a dataset is obtained by averaging over all target agents.

\subsection{Window-Based Metrics}

Other metrics rely on windows with a specific size and orientation for evaluation. A simple example is the miss rate (MR), which counts the proportion of agents that are misses, \ie, where none of the predicted endpoints are within a window $W$ around the ground truth $\bm y_{nT}$. For a single agent $n$ a miss is defined as
\begin{equation}
    \text{Miss}(\bm y_{nT}, \{\hat{\bm x}_{nkT}^\text{pos}\}_{k=1}^K) = \mathbb I(\nexists k:\hat{\bm x}_{nkT}^\text{pos} \in W(\bm y_{nT})),
\end{equation}
where $\mathbb I$ is the indicator function. Argoverse~2 defines the window to be a circle with a radius of \SI{2}{\meter}. Waymo uses a rectangle aligned to the ground-truth heading $\bm y_{nT}^\text{hd}$ whose size depends on the agent's velocity at the last historical timestep.

Average precision (AP) is a window-based metric that also considers confidences (\cref{fig:map}) which are not required to sum to $1$. To compute AP, predictions for all agents across the dataset are sorted based on their confidence. For each confidence level, precision and recall are computed. A prediction counts as a true positive, if its endpoint is the highest-confidence endpoint within the window $W$ around the ground-truth endpoint. Predictions with lower confidences within the window either count as false positive (AP) or are ignored (soft AP). Predictions outside the window are considered false positives and windows not hit by any predictions become false negatives. From these, a precision-recall curve is computed (\cref{fig:smooth}), where standard object detection interpolation is applied \cite{everingham2010pascal}.

To compute mean average precision (mAP), agents are categorized into buckets based on their ground-truth trajectories (\eg, straight, left, stationary). AP is computed independently for each bucket and then averaged. Soft mAP is computed analogously and serves as the main metric for Waymo, using the same window definition as the miss rate.

\PAR{Conflicting objectives.}
Distance-based metrics and window-based metrics reward different behaviors from a forecasting model. Distance-based metrics improve if at least one predicted trajectory is very close to the ground truth. A model is rewarded if it puts many trajectories close together into the main mode, even if they are redundant. By contrast, window-based metrics care about coverage: models score well if they have one prediction in the ground-truth window, additional close-by trajectories either bring no benefit or are actively penalized. This incentivizes a spread between predictions. These goals are inherently conflicting: concentrating predictions to minimize distance helps minFDE but hurts coverage, whereas spreading predictions to cover many windows improves mAP or miss rate but worsens distance-based metrics.

\subsection{Metric-Driven Modeling}
\label{sec:metric_train}

Many current state-of-the-art models implicitly or explicitly tailor their training objectives and post-processing schemes to either distance-based or window-based metrics. Because these metrics reward conflicting behaviors, such metric-dependent training encourages models to specialize for one metric at the expense of others, rather than to learn a generally useful predictive distribution that can support diverse downstream tasks.

Models optimized to train for distance-based metrics often predict exactly $K=6$ trajectories and directly train these outputs by minimizing a distance error between the best trajectory and the ground truth. Concretely, they apply the winner-takes-all (WTA) paradigm which selects the best mode based on Euclidean distance and regress the corresponding trajectory with an $L_1$ loss \cite{lan2024sept}, smooth $L_1$ loss \cite{zhang2024demo, zhang2025polaris}, a Laplacian negative log-likelihood (\ie, an $L_1$ loss with a learned scale) \cite{zhou2022hivt, knoche2025donut, wang2025futurenetlof, zhou2023qcnet}, or a Gaussian negative log-likelihood (\ie, a squared $L_2$ loss with a learned scale) \cite{sun2025impact, shi2022mtr, sun2025rmpyolo}. This procedure mirrors the definition of minFDE, where only the closest prediction contributes to the metric. This improves distance-based metrics but leads to tightly clustered predictions and poor performance on window-based metrics \cite{sun2025impact, shi2022mtr, sun2025rmpyolo}.

Generative, GPT-like trajectory forecasting approaches, which can sample an arbitrary number of trajectories, use $k$-means clustering as a post-processing step to select $K$ trajectories for reporting to the benchmarks \cite{seff2023motionlm, huang2025drivegpt}. Similarly, several works apply $k$-means clustering on ensembles of trajectories to further reduce distance-based errors \cite{zhang2024demo, zhou2023qcnet, zhang2025polaris, wang2025futurenetlof}. In both cases, clustering minimizes the average squared Euclidean distance between sampled trajectories and the chosen cluster centers, which is a direct surrogate for squared minFDE under the model’s predictive distribution. Again, this prioritizes minimizing distance to the ground truth over modeling the full future distribution.

Methods targeting window-based metrics adapt training and post-processing to encourage coverage of many distinct evaluation windows. A common pattern on Waymo is to output $64$ candidate trajectories and then apply non-maximum suppression (NMS) as post-processing \cite{shi2022mtr, shi2024mtrpp, shi2024mtrv3, sun2025rmpyolo, nayakanti2023wayformer}. By suppressing predictions that fall close to each other, NMS decreases the number of trajectory endpoints within a window. This is rewarded by (soft) mAP, where only the highest-confidence hit within each window can contribute to the score. Some methods even make the NMS radius depend on the predicted trajectory length \cite{shi2022mtra, sun2025impact, lin2024eda, yan2025trajflow}, which approximates the velocity-dependent window size used by Waymo and thereby aligns the post-processing even more closely with the specific definition of soft mAP. While these strategies improve mAP and miss rate, they typically worsen minFDE \cite{sun2025impact, shi2022mtr, sun2025rmpyolo}.

ModeSeq \cite{zhou2025modeseq} goes one step further by aligning its training criterion precisely with the window-based metrics. In its early-match-takes-all scheme, a match is defined exactly as a true positive in the benchmark: using a circular window of radius \SI{2}{\meter} for Argoverse~2 and a heading-aligned rectangle whose size depends on the last observed velocity for Waymo. This directly trains the model to avoid predicting multiple trajectories within the same window and tailors the optimization to the precise hit definitions of the metrics. While this improves the targeted window-based scores, it further biases the model toward the benchmark objective and away from modeling the full predictive distribution.

As many current methods are tightly coupled to particular evaluation protocols, instead of aiming for general representations, models cannot be easily applied in novel contexts. In the following, we instead advocate training with metrics-agnostic probabilistic objectives that prioritize the quality and calibration of the predictive distribution. We treat benchmark metrics as downstream problems that are addressed at evaluation time via metric-specific policies.

\section{Metric-Specific Policies}

Instead of adapting the training procedure to improve a certain metric, we propose to decouple the training objective from the evaluation metrics. Concretely, models should be trained to output a calibrated predictive distribution over the agents' future trajectories, which is the most general representation of uncertainty. For evaluation, we propose to continue using existing benchmarks, which require reporting $K$ trajectories with associated confidences $\{(\hat{\bm X}_{nk}, \hat\pi_{nk})\}_{k=1}^K$. To obtain these, we introduce TraDiE policies for Trajectory Distribution Evaluation, which output trajectories and confidences that are optimal for a specific benchmark metric given the predictive distribution. This way, we can assess the quality of the predictive distribution through standard benchmark metrics.

Given a model providing a predictive distribution $p_n(\bm X_n)$ over future trajectories $\bm X_n$, we define a distribution-aware policy as a mapping
\begin{equation}
    \mathcal P: p_n(\bm X_n) \mapsto \{(\hat{\bm X}_{nk}, \hat \pi_{nk})\}_{k=1}^K.
\end{equation}
Ideally, such a policy produces trajectories and confidences that are optimal for a given evaluation metric under the assumption that the ground truth is drawn from the model’s predictive distribution.

In the following, we design policies for the most commonly used trajectory forecasting metrics, which are evaluated at the endpoint of the forecasting horizons. For these, we only require the predictive endpoint distribution $p_{nT}(\bm x_{nT})$, \ie, the marginal of $p_{n}(\bm X_n)$ over the final timestep $T$. The policies then operate on endpoints instead of full trajectories. Depending on the metric, $\bm x_{nT}$ is required to contain either only the positions $\bm x_{nT}^\text{pos}$, or both positions and headings $\bm x_{nT}^\text{hd}$. Implementation details are provided in App.~\ref{app:impl}.

\subsection{Policies for Distance-Based Metrics}

\PARbegin{minFDE.} To obtain the optimal position predictions $\{\hat{\bm x}_{nkT}^\text{pos}\}_{k=1}^K$ for the minFDE metric given the model's predicted distribution $p_{nT}(\bm x_{nT})$, we define the optimization objective for the minFDE policy as
\begin{equation}
    \mathbb E_{\bm x_{nT} \sim p_{nT}} \left[\min_k \|\hat{\bm x}_{nkT}^\text{pos} - \bm x_{nT}^\text{pos}\|_2\right].
    \label{eq:riskfde}
\end{equation}
This is the expected minFDE for the predictions $\{\hat{\bm x}_{nkT}^\text{pos}\}_{k=1}^K$ under the predictive distribution $p_{nT}(\bm x_{nT})$. We can optimize this by using Monte Carlo sampling to obtain a set of positions, $S$, from the endpoint distribution following
\begin{equation}
    S = \{\bm x_{nT}^{(1)}, \dots, \bm x_{nT}^{(M)}\}, \quad \bm x_{nT}^{(m)}\sim p_{nT}(\bm x_{nT}),
    \label{eq:MCs}
\end{equation}
and minimizing the empirical approximation
\begin{equation}
   \frac{1}{|S|} \sum_{\bm x_{nT} \in S} \min_k \|\hat{\bm x}_{nkT}^\text{pos} - \bm x_{nT}^\text{pos}\|_2
    \label{eq:optfde}
\end{equation}
by treating the endpoints $\hat{\bm x}_{nkT}^\text{pos}$ as free variables and optimizing them using gradient descent (\cref{fig:policies}, top).

\subsection{Policies for Window-Based Metrics}
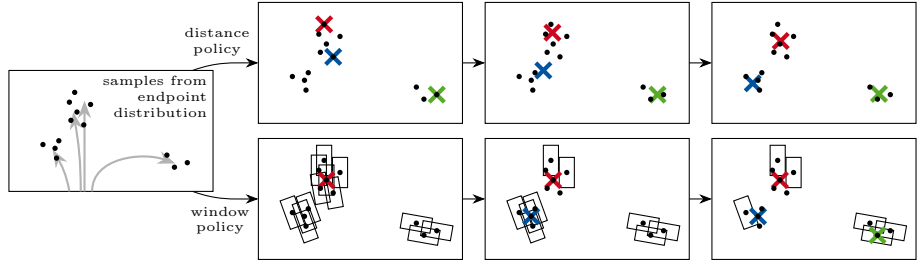
\begin{figure}[t]
  \centering
  \begin{tikzpicture}[font=\footnotesize]

    \def\panelw{2.7}
    \def\panelh{1.6}
    \def\xsep{0.3}
    \def\ysep{0.2}
    \def\windoww{0.2}
    \def\windowh{0.4}
    \def\cl{0.1}

    \def\samplelist{%
      1.08/1.15,
      0.99/0.88,
      0.87/1.31,
      0.82/0.94,
      0.80/1.18,
      0.90/1.05,
      0.61/0.57,
      0.66/0.67,
      0.63/0.44,
      0.45/0.62,
      2.09/0.48,
      2.19/0.32,
      2.36/0.38
    }
    \def\headinglist{%
      180,
      190,
      180,
      190,
      180,
      180,
      200,
      200,
      200,
      200,
      80,
      80,
      80
    }

    \begin{scope}[shift={(0,{0.5*(\panelh+\ysep)})}]
      \draw[black] (0,0) rectangle (\panelw, \panelh);

      \draw[trajectory, thick, black!30] (1, 0) -- (1, 1.2);
      \draw[trajectory, thick, black!30] (0.95, 0) to[out=90, in=280] (0.85, 1.05);
      \draw[trajectory, thick, black!30] (0.8, 0) to[out=100, in=300] (0.58, 0.55);
      \draw[trajectory, thick, black!30] (1.1, 0) to[out=90, in=160] (2.2, 0.37);
      
      \foreach \x/\y in \samplelist {
        \fill[black] (\x,\y) circle (1pt);
      }

    \end{scope}

    \begin{scope}[shift={({\panelw+2*\xsep},{\panelh+\ysep})}]
      \draw[black] (0,0) rectangle (\panelw,\panelh);

      \foreach \x/\y/\col in {%
        0.87/1.31/RWTHRot,
        0.99/0.88/RWTHBlau,
        2.36/0.38/RWTHGruen%
      }{
        \draw[ultra thick,\col]
          (\x-\cl,\y-\cl)--(\x+\cl,\y+\cl);
        \draw[ultra thick,\col]
          (\x-\cl,\y+\cl)--(\x+\cl,\y-\cl);
      }
      \foreach \x/\y in \samplelist {
        \fill[black] (\x,\y) circle (1pt);
      }
    \end{scope}
    
    \begin{scope}[shift={({2*\panelw+3*\xsep},{\panelh+\ysep})}]
      \draw[black] (0,0) rectangle (\panelw,\panelh);

      \foreach \x/\y/\col in {%
        0.89/1.20/RWTHRot,
        0.77/0.70/RWTHBlau,
        2.28/0.38/RWTHGruen%
      }{
        \draw[ultra thick,\col]
          (\x-\cl,\y-\cl)--(\x+\cl,\y+\cl);
        \draw[ultra thick,\col]
          (\x-\cl,\y+\cl)--(\x+\cl,\y-\cl);
      }
      \foreach \x/\y in \samplelist {
        \fill[black] (\x,\y) circle (1pt);
      }
    \end{scope}
    
    \begin{scope}[shift={({3*\panelw+4*\xsep},{\panelh+\ysep})}]
      \draw[black] (0,0) rectangle (\panelw,\panelh);

      \foreach \x/\y/\col in {%
        0.91/1.09/RWTHRot,
        0.54/0.53/RWTHBlau,
        2.21/0.39/RWTHGruen%
      }{
        \draw[ultra thick,\col]
          (\x-\cl,\y-\cl)--(\x+\cl,\y+\cl);
        \draw[ultra thick,\col]
          (\x-\cl,\y+\cl)--(\x+\cl,\y-\cl);
      }
      \foreach \x/\y in \samplelist {
        \fill[black] (\x,\y) circle (1pt);
      }
    \end{scope}

    \begin{scope}[shift={({\panelw+2*\xsep},0)}]
      \draw[black] (0,0) rectangle (\panelw,\panelh);

      \def\visiblelist{1, 1, 1, 1, 1, 1, 1, 1, 1, 1, 1, 1, 1}
      
      \foreach[count=\i] \x/\y in \samplelist {
        \pgfmathparse{{\headinglist}[\i-1]}
        \let\heading\pgfmathresult
    
        \pgfmathparse{{\visiblelist}[\i-1]}
        \let\visible\pgfmathresult
    
        \ifdim\visible pt>0pt
          \draw[black, rotate around={\heading:(\x,\y)}]
            (\x-.5*\windoww,\y-.5*\windowh) rectangle
            (\x+.5*\windoww,\y+.5*\windowh);
        \fi
      }
      
      \foreach \x/\y/\col in {%
        0.90/1.05/RWTHRot%
      }{
        \draw[ultra thick,\col]
          (\x-\cl,\y-\cl)--(\x+\cl,\y+\cl);
        \draw[ultra thick,\col]
          (\x-\cl,\y+\cl)--(\x+\cl,\y-\cl);
      }

      \foreach \x/\y in \samplelist {
        \fill[black] (\x,\y) circle (1pt);
      }
      
    \end{scope}
    
    \begin{scope}[shift={({2*\panelw+3*\xsep},0)}]
      \draw[black] (0,0) rectangle (\panelw,\panelh);

      \def\visiblelist{1, 0, 1, 0, 0, 0, 1, 1, 1, 1, 1, 1, 1}
    
      \foreach[count=\i] \x/\y in \samplelist {
        \pgfmathparse{{\headinglist}[\i-1]}
        \let\heading\pgfmathresult
    
        \pgfmathparse{{\visiblelist}[\i-1]}
        \let\visible\pgfmathresult
    
        \ifdim\visible pt>0pt
          \draw[black, rotate around={\heading:(\x,\y)}]
            (\x-.5*\windoww,\y-.5*\windowh) rectangle
            (\x+.5*\windoww,\y+.5*\windowh);
        \fi
      }
      
      \foreach \x/\y/\col in {%
        0.90/1.05/RWTHRot,
        0.61/0.57/RWTHBlau%
      }{
        \draw[ultra thick,\col]
          (\x-\cl,\y-\cl)--(\x+\cl,\y+\cl);
        \draw[ultra thick,\col]
          (\x-\cl,\y+\cl)--(\x+\cl,\y-\cl);
      }

      \foreach \x/\y in \samplelist {
        \fill[black] (\x,\y) circle (1pt);
      }
    \end{scope}
    
    \begin{scope}[shift={({3*\panelw+4*\xsep},0)}]
      \draw[black] (0,0) rectangle (\panelw,\panelh);

      \def\visiblelist{1, 0, 1, 0, 0, 0, 0, 0, 0, 1, 1, 1, 1}
    
      \foreach[count=\i] \x/\y in \samplelist {
        \pgfmathparse{{\headinglist}[\i-1]}
        \let\heading\pgfmathresult
    
        \pgfmathparse{{\visiblelist}[\i-1]}
        \let\visible\pgfmathresult
    
        \ifdim\visible pt>0pt
          \draw[black, rotate around={\heading:(\x,\y)}]
            (\x-.5*\windoww,\y-.5*\windowh) rectangle
            (\x+.5*\windoww,\y+.5*\windowh);
        \fi
      }
      
      \foreach \x/\y/\col in {%
        0.90/1.05/RWTHRot,
        0.61/0.57/RWTHBlau,
        2.19/0.32/RWTHGruen%
      }{
        \draw[ultra thick,\col]
          (\x-\cl,\y-\cl)--(\x+\cl,\y+\cl);
        \draw[ultra thick,\col]
          (\x-\cl,\y+\cl)--(\x+\cl,\y-\cl);
      }

      \foreach \x/\y in \samplelist {
        \fill[black] (\x,\y) circle (1pt);
      }
    \end{scope}
    \draw[trajectory] (0.9*\panelw, {1.5*\panelh+0.5*\ysep}) to[out=20, in=180] node[anchor=east, black, pos=1, yshift=8pt,align=center,font=\tiny] {distance\\policy} (\panelw+2*\xsep, {1.5*\panelh+\ysep});
    \draw[trajectory] (0.9*\panelw, {0.5*\panelh+0.5*\ysep}) to[out=-20, in=180] node[anchor=east, black, pos=1, yshift=-8pt,align=center,font=\tiny] {window\\policy} (\panelw+2*\xsep, {0.5*\panelh});
    \draw[trajectory] ({2*\panelw+2*\xsep}, {1.5*\panelh+\ysep}) -- ({2*\panelw+3*\xsep}, {1.5*\panelh+\ysep});
    \draw[trajectory] ({3*\panelw+3*\xsep}, {1.5*\panelh+\ysep}) -- ({3*\panelw+4*\xsep}, {1.5*\panelh+\ysep});
    \draw[trajectory] ({2*\panelw+2*\xsep}, {0.5*\panelh}) -- ({2*\panelw+3*\xsep}, {0.5*\panelh});
    \draw[trajectory] ({3*\panelw+3*\xsep}, {0.5*\panelh}) -- ({3*\panelw+4*\xsep}, {0.5*\panelh});

    \node[anchor=north east, align=right, font=\tiny, inner sep=2pt] at (\panelw, 1.5*\panelh+0.5*\ysep) {samples from\\endpoint\\distribution};
    
  \end{tikzpicture}
  \caption{\textbf{Illustration of two policies.} Samples from the endpoint distribution (\textbullet) of the \textbf{\color{black!50} prediction} (left) are used to optimize three endpoints (\textbf{\color{RWTHRot}\texttimes}, \textbf{\color{RWTHBlau}\texttimes}, \textbf{\color{RWTHGruen}\texttimes}). The top row shows gradient descent for minFDE; the bottom row shows the iterative process of choosing the sample covered by the most rectangles for soft mAP.}
  \label{fig:policies}
\end{figure}

\PARbegin{(Soft) mAP.} Computing an optimal policy for average precision is challenging, because it depends on the global ranking of all predictions across all agents and scenes. To obtain a tractable approximation, we design a per-agent heuristic. We first draw a set of candidate endpoints from the predictive distribution and associate each candidate with an evaluation window. We then greedily select endpoints that are covered by the largest number of windows, while ignoring windows that have already been hit in previous steps.

More formally, let $W_n({\bm y}_{nT})$ denote the window defining a \emph{hit} centered at the ground-truth endpoint ${\bm y}_{nT}^\text{pos}$ with orientation ${\bm y}_{nT}^\text{hd}$. A specific predicted endpoint $\hat{\bm x}_{nkT}^\text{pos}$ from the set of predictions $\{(\hat{\bm x}_{nkT}^\text{pos}, \hat \pi_{nk})\}_{k=1}^K$ with confidences $\hat \pi_{n1} \geq \hat \pi_{n2} \geq \dots \geq \hat \pi_{nK}$ is counted as a true positive ($\text{TP}$) if it lies within the window $W_n({\bm y}_{nT})$ and no prediction with
higher confidence lies in the same window. Formally, we define
\begin{equation}
    \text{TP}_{nk}(\hat{\bm x}_{nkT}^\text{pos}, \bm y_{nT}) = \mathbb I\left(\hat{\bm x}_{nkT}^\text{pos} \in W_n(\bm y_{nT}) \land \forall i<k: \hat{\bm x}_{niT}^\text{pos} \notin W_n(\bm y_{nT})\right).
     \label{eq:tp}
\end{equation}
As the ground-truth endpoint $\bm y_{nT}$ is not available at inference time, we use samples $\tilde{\bm x}_{nT}$ from the predictive distribution $p_{nT}$ instead. The probability that $\hat{\bm x}_{nkT}$ is the true positive under the predictive distribution is then given by
\begin{equation}
    \overline{\text{TP}}_{nk}(\hat{\bm x}_{nkT}^\text{pos}) = \mathbb E_{\tilde{\bm x}_{nT} \sim p_{nT}} \left[ \text{TP}_{nk}(\hat{\bm x}_{nkT}^\text{pos}, \tilde{\bm x}_{nT})\right].
\end{equation}

In practice, we approximate this expectation by averaging over samples $\tilde{\bm x}_{nT}$ of a Monte Carlo set $S$ (\cref{fig:policies}, bottom). To find predictions $\{\hat{\bm x}_{nkT}^\text{pos}\}_{k=1}^K$, a greedy heuristic iteratively picks $\hat{\bm x}_{nkT}^\text{pos}$ for $k\in\{1,\dots,K\}$ from the same set $S$ with
\begin{equation}
    \hat{\bm x}_{nkT}^\text{pos} = \argmax_{\bm x_{nT} \in S} \overline{\text{TP}}_{nk}(\bm x_{nT}^\text{pos}),
\end{equation}
\ie, we choose the endpoint to which we assign the highest probability of becoming a true positive. We define the corresponding confidences as
\begin{equation}
    \hat{\pi}_{nk} = \overline{\text{TP}}_{nk}(\hat{\bm x}_{nkT}^\text{pos}).
\end{equation}

This approach has several properties that naturally align with improving average precision. By repeatedly choosing the endpoint that is most likely to be the highest-confidence true positive, we increase the chance that true positives appear earlier than false positives in the global ranking, which improves precision across confidence thresholds. The greedy selection is also a standard heuristic \cite{hochbaum1997setcover} for the maximum coverage problem: prioritizing endpoints that cover many yet-uncovered windows reduces the number of false negatives and thus improves recall and AP. Finally, because windows that have already been hit are ignored in subsequent steps, the procedure avoids placing multiple predictions in the same window, which would either be ignored (soft mAP) or penalized (mAP).

\PAR{Miss rate.} A forecast is considered a miss if none of the $K$ predictions are within a window around the target. The miss rate is the proportion of misses across the dataset. Argoverse~2 defines this region as a circle with a radius of \SI{2}{\meter}; Waymo uses the same definition as for the (soft) mAP window. Confidences do not matter for this metric. For an agent $n$, the probability of a hit (at least one prediction in the window) under the predictive distribution is
\begin{equation}
    \sum_{k=1}^K\overline{\text{TP}}_{nk}(\hat{\bm x}_{nkT}^\text{pos}).
    \label{eq:optmr}
\end{equation}
Thus, this metric is optimized when we select $\{\hat{\bm x}_{nkT}^\text{pos}\}_{k=1}^K$ to maximize \cref{eq:optmr}, \ie, we aim to maximize the total number of covered windows under the predictive distribution. As our greedy heuristic for (soft) mAP aims to optimize this objective, the same heuristic policy is applicable to soft mAP, mAP, and miss rate.

\section{DONUT-NLL}

\subsection{Baseline}

To study metric-agnostic training and metric-specific evaluation policies in a controlled setting, we instantiate our approach on top of DONUT \cite{knoche2025donut}, a decoder-only transformer for trajectory forecasting that obtains state-of-the-art results on Argoverse~2 \cite{wilson2021argoverse2}. For each agent $n \in \{1, \dots, N\}$ and each predicted trajectory $k \in \{1, \dots, K\}$, DONUT predicts a $T$-step future trajectory. It parameterizes positions with 1D Laplace distributions over the global $x$- and $y$-axes by predicting $\bm \mu_{n}^x, \bm \mu_{n}^y, \bm b_{n}^x, \bm b_{n}^y \in \mathbb R^{T \times K}$, and it models heading with von Mises distributions with parameters $\bm \mu_{n}^\text{hd}, \bm \kappa_{n}^\text{hd} \in \mathbb R^{T \times K}$. In addition, mixture weights $\pi_n \in \mathbb R^K$ are obtained from the decoder via a small prediction head, representing non-negative mode probabilities over the $K$ trajectories.

To better align the positional uncertainty with the kinematics of the agents, we make the following adaptation to DONUT: we use the predicted mean heading $\bm \mu_{nkt}^\text{hd}$ to align the 1D distributions with the agent's local longitudinal (lg) and lateral (lt) directions, instead of aligning with global $x$- and $y$-axes. Denoting the ground-truth observation at timestep $t$ for agent $n$ by $\bm y_{nt} = (\bm y_{nt}^\text{lg}, \bm y_{nt}^\text{lt}, \bm y_{nt}^\text{hd})$, the likelihood at each timestep for a given agent and mode can be computed as
\begin{equation}
    \hspace{-0.11cm}
    p_{nkt}(\bm y_{nt}) = \text{L}(\bm y_{nt}^\text{lg} \mid \bm \mu_{nkt}^\text{lg}, \bm b_{nkt}^\text{lg}) \cdot \text{L}(\bm y_{nt}^\text{lt} \mid \bm \mu_{nkt}^\text{lt}, \bm b_{nkt}^\text{lt}) \cdot \text{vM}(\bm y_{nt}^\text{hd} \mid \bm \mu_{nkt}^\text{hd}, \bm \kappa_{nkt}^\text{hd})
    \label{eq:pertimestep}
\end{equation}
with $\text{L}(\cdot)$ and $\text{vM}(\cdot)$ denoting 1D Laplace and von Mises densities, respectively.

To optimize the mixture components for the distance-based metrics of the Argoverse~2 benchmark, DONUT uses the winner-takes-all paradigm and selects the best mode $k_n^*$ using the average Euclidean distance over all timesteps. This component's log-likelihood is optimized via
\begin{equation}
    \mathcal L_\text{reg} = -\sum_n \sum_{t=1}^T \log p_{nk_n^*t}(\bm y_{nt}).
\end{equation}
A classification loss optimizes the log-likelihood of the mixture components' weights at the final timestep $T$, with $\bar p$ indicating a no-gradient version of the density:
\begin{equation}
    \mathcal L_\text{cls} = -\sum_n \log \sum_{k=1}^K \pi_{nk} \bar p_{nkT}(\bm y_{nT}).
\end{equation}

\subsection{Probabilistic Training Objectives}

To match our paradigm of training a model to directly optimize a predictive distribution rather than specific metrics, we adjust DONUT and propose two new variants: Traj-NLL and Step-NLL. Instead of using the winner-takes-all objective, Traj-NLL directly optimizes the predictive distribution with the per-timestep likelihood in \cref{eq:pertimestep} and minimizes the negative log-likelihood (NLL):
\begin{equation}
    -\log p(\bm Y) = -\sum_n \log \sum_{k=1}^K \pi_{nk} \prod_{t=1}^T p_{nkt}(\bm y_{nt}).
\end{equation}
In this formulation, the mixture weights $\pi_{nk}$ are shared across all timesteps $t$, so each component $k$ corresponds to a complete trajectory hypothesis.

To increase temporal flexibility, we further propose Step-NLL that defines a mixture for each timestep independently, yielding the negative log-likelihood
\begin{equation}
    -\log p(\bm Y) = -\sum_n \sum_{t=1}^T \log \sum_{k=1}^K \pi_{nkt} \, p_{nkt}(\bm y_{nt}).
\end{equation}
In this second variant, we introduce time-dependent mixture weights $\pi_{nkt}$, so that the importance of each component is allowed to vary across timesteps. This increases the expressiveness of the model, as different modes can become more or less relevant over time, but it also weakens the implicit coupling between timesteps that is present when a single $\pi_{nk}$ is shared across the entire horizon.

\PAR{Position distributions.}
Using our policies, we systematically investigate different position distribution families as drop-in replacements for DONUT’s original Laplace distribution. A common choice \cite{sun2025impact, sun2025rmpyolo, shi2024mtrv3} is the Gaussian distribution. Compared to Laplace, it has lighter tails and a less sharp peak. In preliminary experiments, Gaussians performed poorly, which is why we experiment with two generalizations of the Gaussian instead.

The generalized Gaussian distribution includes an additional parameter that controls the density's shape. Laplace and Gaussian distributions are special cases of the generalized Gaussian. We further consider discrete scale mixtures of Gaussians with a fixed number of $J=5$ components that share a single mean but differ in scale. We provide more details in App.~\ref{app:distributions}.

\section{Experiments}

\PAR{Dataset.} We evaluate on the large-scale Waymo motion prediction benchmark \cite{ettinger2021waymo}, which reports a number of window-based and distance-based metrics. It consists of approximately 487k training scenes, with 44k scenes for evaluation and 45k scenes for testing. Models must predict $K=6$ future trajectories over \SI{8}{\second} with assigned confidences, using \SI{1.1}{\second} of historical input, sampled at \SI{10}{\hertz}.

\PAR{Additional baselines.}
To assess transferability beyond DONUT-NLL, we apply our metric-specific policies to the open-source trajectory forecasting models MTR~\cite{shi2022mtr} and QCNet~\cite{zhou2023qcnet}. MTR was originally evaluated on Waymo and is therefore optimized for window-based metrics. It predicts a $64$-component Gaussian mixture and uses non-maximum suppression to select $6$ trajectories. We report its original NMS outputs as the naive baseline and additionally evaluate our policies by sampling from the full $64$-component mixture. By contrast, QCNet targets Argoverse 2 and directly outputs $6$ trajectory modes parameterized as Laplacians. For both models, we vary the training objective and compare the original WTA losses to our probabilistic objectives (Traj-NLL and Step-NLL).

\PAR{Main experiments.} Our experiments are designed to answer six main questions: (i) are our TraDiE policies effective at turning a single distribution into strong performance across heterogeneous metrics, (ii) do our policies detect poorly calibrated predictive distributions, (iii) how does the training objective (WTA, Traj-NLL, Step-NLL) affect the quality of the learned predictive distribution, (iv) does optimizing for distributions instead of metrics favor different design choices, (v) which position distributions are best suited for our paradigm, and (vi) is the analysis of predictive distributions via policies model-agnostic?

\begin{table}[t]
    \centering
    \smalltext
    \renewcommand{\tabcolsep}{2pt}
    \caption{\textbf{Study of training objectives with Laplace distributions.} Applying our distance or window policies (indicated by $\rightarrow$) combined with NLL training consistently outperforms the naive WTA predictions. Notably, naive evaluation suggests different design choices than evaluation with policies (\bestnaive{\phantom{X}} vs.\ \bestpolicy{\phantom{X}}). Evaluation on Waymo~\cite{ettinger2021waymo} \textit{val}.} 
    \label{tab:loss_ablation}
    \begin{tabularx}{\linewidth}{lYYYY} \toprule
        Objective      & Soft mAP \up & mAP \up & Miss rate \down &  minFDE \down \\
        \midrule
        WTA      & \policyarrow{0.3427}{\textbf{0.4764}} & \policyarrow{0.2995}{\textbf{0.4736}} & \policyarrow{0.1310}{\textbf{0.1044}} & \policyarrow{\bestnaive{\textbf{1.0520}}}{1.0824} \\
        Traj-NLL & \policyarrow{0.3401}{\textbf{0.4921}} & \policyarrow{0.2851}{\textbf{0.4892}} & \policyarrow{\bestnaive{0.1286}}{\textbf{0.0946}} & \policyarrow{1.0874}{\textbf{1.0374}} \\
        Step-NLL &\policyarrow{ \bestnaive{0.3595}}{\bestpolicy{\textbf{0.5082}}} & \policyarrow{\bestnaive{0.3163}}{\bestpolicy{\textbf{0.5053}}} & \policyarrow{0.1444}{\bestpolicy{\textbf{0.0902}}} & \policyarrow{1.2236}{\bestpolicy{\textbf{1.0200}}} \\
        \bottomrule
    \end{tabularx}
    \label{tab:policy}
\end{table}

In \cref{tab:loss_ablation}, we report results for the WTA, Traj-NLL, and Step-NLL training objectives, both before and after our metric-specific policies. For both NLL variants, which optimize the predictive distribution directly, we find that our policies greatly improve the corresponding metrics. The distance policy consistently reduces minFDE, and the window policy leads to substantial improvements in soft mAP, mAP, and miss rate. This confirms the first key aspect of our paradigm: given a single predictive distribution, the proposed policies can effectively optimize different metrics without retraining the underlying model.

Applying the distance policy to WTA degrades minFDE compared with naive evaluation, since WTA directly optimizes minFDE rather than a calibrated distribution. The result is therefore expected: if the predictive distribution poorly aligns with observed uncertainty, optimizing the metric under that distribution may not yield good results. This suggests that our policies expose miscalibration rather than merely improving benchmark scores through post-processing.

Comparing the different training objectives, we observe that Step-NLL consistently outperforms both WTA and Traj-NLL once TraDiE policies are applied. This directly supports our central hypothesis: optimizing the predictive distribution directly, via NLL, is more beneficial than optimizing a surrogate metric-specific loss once metric-specific policies are used at evaluation time.

We can further observe that design decisions differ between models trained to optimize metrics and models trained for distributions: without policies, the best objective for minFDE is WTA, which directly optimizes the distance-based loss. By contrast, Step-NLL outputs the best predictive distribution, as it consistently outperforms the other objectives if we apply our policies.

\begin{table*}[t]
    \centering
    \smalltext
    \renewcommand{\tabcolsep}{2pt}
    \caption{\textbf{Study of position distributions using Step-NLL.} Generalized Gaussians outperform the alternatives with policy optimization. Evaluation on Waymo~\cite{ettinger2021waymo} \textit{val}.}
    \label{tab:distributions}
    \begin{tabularx}{1.0\linewidth}{lYYYY} \toprule
        Position Distr. & Soft mAP \up & mAP \up & Miss rate \down & minFDE \down \\
        \midrule
        Laplace         & \policyarrow{0.3595}{\best{0.5082}} & \policyarrow{0.3163}{\best{0.5053}} & \policyarrow{\bestnaive{0.1444}}{\best{0.0902}} & \policyarrow{\bestnaive{1.2236}}{\best{1.0200}}\\
        Scale Mixture   & \policyarrow{\bestnaive{0.3837}}{\best{0.5053}} & \policyarrow{\bestnaive{0.3533}}{\best{0.5024}} & \policyarrow{0.1492}{\best{0.0909}} & \policyarrow{1.2722}{\best{1.0218}}\\
        Gen.\ Gaussian  & \policyarrow{0.3759}{\bestpolicy{\best{0.5101}}} & \policyarrow{0.3439}{\bestpolicy{\best{0.5070}}} & \policyarrow{0.1561}{\bestpolicy{\best{0.0876}}} & \policyarrow{1.3127}{\bestpolicy{\best{1.0103}}}\\
        \bottomrule
    \end{tabularx}
\end{table*}

In \cref{tab:distributions}, we investigate how the choice of position distribution influences performance under our paradigm. We compare Laplace, generalized Gaussian, and Gaussian scale mixture distributions, each combined with Step-NLL and evaluate them naively and with our metric-specific policies. Once our policies are applied, the generalized Gaussian distribution consistently outperforms the alternatives across all metrics. This suggests that its additional shape flexibility over the Laplacian is beneficial for learning a well-calibrated predictive distribution that can be effectively exploited by the downstream policy.

As before, the naive evaluation again favors different design choices. For minFDE, naive Laplace performs best, likely because its sharper shape encourages the model to place modes close to the ground-truth endpoint. In contrast, naive Gaussian scale mixtures perform best for soft mAP and mAP, presumably because their smoother and broader predictive distributions improve coverage. The consistent advantage of the generalized Gaussian after policy optimization indicates that it learns a better-calibrated predictive distribution. We provide full ablations over all combinations of losses and distribution families in App.~\ref{app:full_abl}.

\begin{table}[t]
    \centering
    \smalltext
    \renewcommand{\tabcolsep}{2pt}
    \caption{\textbf{Additional baselines}. MTR \cite{shi2022mtr} consistently improves; QCNet \cite{zhou2023qcnet} degrades for minFDE due to its poor predictive distributions. Evaluation on Waymo \emph{val}.}
    \label{tab:mtrqcnet}
    \begin{tabularx}{1.0\linewidth}{llYYYY}
    \toprule
        \multicolumn{2}{l}{Model} & Soft mAP \up & mAP \up & Miss rate \down & minFDE \down  \\
    \midrule
        \multirow{3}{*}{\rotatebox{90}{MTR}}
          & WTA      & \policyarrow{0.4288}{\best{0.4837}} & \policyarrow{0.4154}{\best{0.4798}} & \policyarrow{0.1386}{\best{0.0979}} & \policyarrow{1.2357}{\best{1.0615}} \\
          & Traj-NLL & \policyarrow{0.3170}{\best{0.4422}} & \policyarrow{0.2849}{\best{0.4395}} & \policyarrow{0.1739}{\best{0.0984}} & \policyarrow{1.5428}{\best{1.0656}} \\
          & Step-NLL & \policyarrow{0.2550}{\best{0.4873}} & \policyarrow{0.2428}{\best{0.4838}} & \policyarrow{0.1938}{\best{0.0924}} & \policyarrow{1.5520}{\best{1.0675}} \\
    \midrule
        \multirow{3}{*}{\rotatebox{90}{QCNet}}
          & WTA      & \policyarrow{0.3236}{\best{0.4045}} & \policyarrow{0.2912}{\best{0.4018}} & \policyarrow{0.1628}{\best{0.1276}} & \policyarrow{\best{1.2254}}{2.2805} \\
          & Traj-NLL & \policyarrow{0.2980}{\best{0.3814}} & \policyarrow{0.2475}{\best{0.3791}} & \policyarrow{0.1743}{\best{0.1343}} & \policyarrow{\best{1.3416}}{2.9618} \\
          & Step-NLL & \policyarrow{0.2401}{\best{0.4209}} & \policyarrow{0.1964}{\best{0.4184}} & \policyarrow{0.1922}{\best{0.1216}} & \policyarrow{\best{1.4207}}{3.1753} \\
    \bottomrule
    \end{tabularx}
\end{table}

\PAR{Additional baselines.} To evaluate whether our findings extend to other models, we apply our policies to MTR \cite{shi2022mtr} and QCNet \cite{zhou2023qcnet} and report the results in \cref{tab:mtrqcnet}. MTR improves consistently, and even outperforms the original post-processing, indicating that the distribution-aware policies transfer to a different architecture. Changing the training loss only has a minor impact in this setup.

For QCNet, minFDE degrades significantly under the distance policy. This suggests that QCNet has not learned a well-calibrated predictive distribution. Our policies directly optimize the expectation of the metric under the model's output distribution, so if there is any discrepancy with the real-world uncertainty, the policy selects trajectories that are suboptimal in reality. We attribute this mismatch to QCNet's position distribution parameterization: its Laplacians are oriented according to the last historical heading, and therefore become misaligned with the future motion direction on curved trajectories. To compensate, QCNet inflates the uncertainty into a more circular shape, which makes the minFDE-optimal endpoints spread out. Window-based metrics are less affected because the distribution’s main mass stays concentrated around each mode.

This is exactly what our policies are meant to expose: if the predictive distribution is miscalibrated, policy optimization cannot find good trajectories for real data. In this sense, policy-optimized evaluation enables a direct diagnostic of distribution quality that naive metric computation can obscure.

\begin{table*}[t]
    \newcommand{\ela}[1]{{\hide{\textsc{#1}}}}
    \centering
    \renewcommand{\tabcolsep}{5pt}
    \caption{\textbf{Comparison to state of the art on Waymo~\cite{ettinger2021waymo} \textit{test}.} We mark ensembles with \ela{e}, and omit methods that train on additional data or use LiDAR.}
    \label{tab:sota_comparison}
    \small
    \begin{tabularx}{\linewidth}{lYYYY} \toprule
        Model & Soft mAP \up & mAP \up & Miss rate \down & minFDE \down \\
        \midrule
        DriveGPT \cite{huang2025drivegpt}         & 0.3795 & ---      & 0.1236 & 1.0609 \\
        RMP-YOLO (e2e) \cite{sun2025rmpyolo}      & 0.3828 & 0.3440 & 0.1354 & 1.0932 \\
        IMPACT (e2e) \cite{sun2025impact}         & 0.4434 & 0.4253 & 0.1274 & \secn{1.0497} \\
        ModeSeq \cite{zhou2025modeseq}            & 0.4487 & 0.4450 & 0.1244 & 1.0836 \\
        EDA \cite{lin2024eda}                     & 0.4510 & 0.4401 & 0.1169 & 1.1702 \\
        UniMotion \cite{song2025unimotion}        & 0.4642 & 0.4534 & 0.1162 & 1.1643 \\
        RMP-YOLO \cite{sun2025rmpyolo}            & 0.4673 & 0.4523 & 0.1160 & 1.1697 \\
        TrajFlow \cite{yan2025trajflow}           & 0.4710 & 0.4604 & 0.1162 & 1.1667 \\
        IMPACT \cite{sun2025impact}               & 0.4721 & 0.4609 & 0.1143 & 1.1540 \\
        ModeSeq \cite{zhou2025modeseq} \ela{e}    & 0.4737 & \secn{0.4665} & 0.1204 & 1.1766 \\
        RMP-YOLO \cite{sun2025rmpyolo} \ela{e}    & 0.4737 & 0.4531 & \underline{0.1084} & 1.1188 \\
        IMPACT \cite{sun2025impact} \ela{e}       & \secn{0.4801} & 0.4598 & 0.1087 & 1.1295 \\
        \midrule
        \multirow{2}{*}{DONUT-NLL} 
                                        \hfill \quad window  & \best{0.5018} & \best{0.4987} & \best{0.0900} & \hide{1.3280}\\
                                        \hfill \quad distance & \hide{0.1649} & \hide{0.1135} & \hide{0.1199} & \best{1.0304} \\
        \bottomrule
    \end{tabularx}
\end{table*}

\PAR{Comparison to the state of the art.}
In \cref{tab:sota_comparison}, we apply both of our metric-specific policies to DONUT-NLL with the Step-NLL loss and the generalized Gaussian position distribution, and compare with state-of-the-art methods. We find that our policies allow a single model to obtain state-of-the-art performance across both distance-based and window-based metrics with the same model weights, whereas existing methods typically perform well for only one of these when naively evaluated. DONUT-NLL outperforms IMPACT (e2e) \cite{sun2025impact} on minFDE, while using roughly $5\times$ fewer parameters. For soft mAP, mAP, and miss rate, it even outperforms ensemble methods while using only a single model. Together, these results validate our proposed paradigm: training a metric-agnostic predictive distribution with probabilistic objectives is an effective way to model future trajectories, and the quality of the predictive distribution can be evaluated on benchmarks using our metric-specific policies.

\section{Limitations and Discussion}

Our TraDiE policies provide a mechanism for adapting general-purpose trajectory prediction models that are trained to output calibrated predictive distributions to common trajectory forecasting metrics such as minFDE and soft mAP. This allows a single predictive model to be flexibly adapted to different downstream objectives without retraining. The main drawback is the additional computational cost incurred at inference time. Because our policies are intended for evaluation and not for direct deployment on autonomous vehicles, we did not focus on optimizing efficiency. In practical settings, however, an explicit trade-off between output quality and computational efficiency would be required.

Another limitation is that our results are restricted to Waymo. Preliminary experiments on Argoverse~2 suggest that Step-NLL is prone to overfitting, possibly because the step-wise mixture weights introduce additional modeling flexibility. Nevertheless, on Waymo, DONUT-NLL combined with our TraDiE policies is the only method to date that reaches state-of-the-art performance on both window-based and distance-based metrics without requiring retraining.

\section{Conclusion}

In this paper, we propose to shift the focus of trajectory forecasting from metric-specific training to learning well-calibrated predictive distributions and treating benchmark metrics as downstream tasks via metric-specific policies. Concretely, we design sampling-based policies for minFDE, miss rate, and (soft) mAP. We instantiate this framework with DONUT-NLL, a negative log-likelihood variant of DONUT. Our experiments on the Waymo motion prediction benchmark confirm that metric-agnostic probabilistic training, combined with our policies, yields a single model that can be adapted post hoc to diverse metrics while surpassing state-of-the-art results. Importantly, this distribution-centric view changes the preferred design choices: DONUT-NLL outperforms WTA, i.e., metric-specific training. This indicates that centering trajectory forecasting on predictive distributions fundamentally changes how forecasting models should be designed. Extending our metric-specific policies to multi-agent forecasting and implementing continuity-aware objectives are interesting directions for future work.

\PAR{Acknowledgments.}
This work was partially funded by the BMBF project 6GEM (16KISK036K).
The authors gratefully acknowledge the computing time provided to them at the NHR Center NHR4CES at RWTH Aachen University (project number p0024673). This is funded by the Federal Ministry of Research, Technology and Space, and the state governments participating on the basis of the resolutions of the GWK for national high performance computing at universities (\httpsurl{www.nhr-verein.de/unsere-partner}).

\bibliographystyle{splncs04}
\bibliography{main}

\newpage
\begin{center}
\Large \textbf{Towards Metric-Agnostic Trajectory Forecasting}\\[1em]
\large \textbf{Supplementary Material}
\end{center}

\appendix

\renewcommand*{\theHsection}{appendix.\Alph{section}}
\renewcommand*{\theHsubsection}{appendix.\Alph{section}.\arabic{subsection}}
\renewcommand*{\theHsubsubsection}{appendix.\Alph{section}.\arabic{subsection}.\arabic{subsubsection}}

\section{Additional Proofs}
\label{app:proofs}

\section{Comparison to HOME \cite{gilles2021home}}

\label{app:home}

In the following, we discuss the similarities and differences between our approach and HOME \cite{gilles2021home} in more detail. Conceptually, both approaches share the idea of first learning a distributional representation over endpoints and then applying metric-specific optimization to extract a finite set of trajectories for evaluation. However, HOME relies on heuristics tailored to improve the empirical performance of its specific model. By contrast, our goal is to define \emph{model-agnostic} policies that can be applied to any probabilistic forecaster and that correspond as closely as possible to optimizing the \emph{exact} benchmark metrics under the predictive distribution.

HOME subdivides trajectory forecasting into three steps: (1) learn an endpoint heatmap from history and scene information, (2) optimize for different metrics within the heatmap, and (3) apply a second network to connect the endpoint and the last historical point to retrieve a full trajectory.

The heatmap is rasterized using square cells with a side length of \SI{0.5}{\meter}. The target heatmap $\bm M$ defines a circular Gaussian around the observed endpoint with a standard deviation of \SI{2}{\meter}. To train the predicted heatmap $\hat{\bm M}$, they weight a variant of the focal loss with a squared error for each cell $i$:
\begin{equation}
    \mathcal L_\text{HOME}(\bm M_i, \hat{\bm M}_i)
    = -(\bm M_i - \hat{\bm M}_i)^2 \, f(\bm M_i, \hat{\bm M}_i),
\end{equation}
with the focal loss variant
\begin{equation}
    f(\bm M_i, \hat{\bm M}_i) =
    \begin{cases}
    \log(\hat{\bm M}_i) & \text{if } \bm M_i = 1,\\
    (1-\bm M_i)^4 \log(1-\hat{\bm M}_i) & \text{otherwise}.
    \end{cases}
\end{equation}
This objective does not produce a normalized, calibrated probability distribution. In contrast, we train models directly with a negative log-likelihood, yielding a normalized predictive distribution that is more suitable for applying policies that optimize evaluation metrics under this distribution.

\PAR{Miss Rate.}
As defined in Eq.~(2), a miss for an agent $n$ occurs when none of the $K$ predicted endpoints lies inside the evaluation window $W(\bm y_{nT})$ around the ground-truth endpoint $\bm y_{nT}$. Given a predictive endpoint distribution $p_{nT}$, the optimal policy for miss rate is therefore to choose $K$ endpoints that maximize the probability of at least one hit, which is exactly the objective formalized in Eq.~(11).

HOME instead applies a greedy algorithm that selects endpoints such that the window around each selected endpoint covers as much probability mass as possible. Note that this is a different objective: instead of finding endpoints covering as many windows as possible, they search for windows covering as many endpoints as possible. For Argoverse~2 \cite{wilson2021argoverse2}, where all windows are circular with fixed radius, both formulations coincide. However, their method does not match the metric definition when windows depend on heading or velocity, as is the case for the rectangular, heading-aligned windows used in Waymo \cite{ettinger2021waymo}. In contrast, our policies are derived directly from the miss rate objective and therefore apply to arbitrary window geometries.

Moreover, HOME does not use the exact benchmark window parameters from Argoverse~2 \cite{wilson2021argoverse2}: instead of performing the optimization with circular windows of radius \SI{2}{\meter}, they use \SI{1.8}{\meter} as this empirically lowers their miss rate. While this is a reasonable engineering choice for maximizing benchmark performance of a particular model, the algorithm does not optimize the miss-rate metric under the learned distribution as defined by the benchmark. In our framework, the policy is part of the evaluation protocol: we commit to using the exact metric definition without additional tuning, so that differences in benchmark scores can be attributed to the quality of the predictive distribution rather than to metric-specific hyperparameters embedded in the policy.

\PAR{minFDE.}
HOME proposes a custom iterative algorithm derived from $k$-means to optimize minFDE. Standard $k$-means minimizes the sum of \emph{squared} Euclidean distances between points and their assigned cluster centers. In contrast, optimizing minFDE requires minimizing the \emph{non-squared} Euclidean distance, so $k$-means does not directly optimize minFDE under the predictive distribution. HOME further modifies the centroid update as
\begin{equation}
    c_k = \frac{1}{N_k} \sum_{i}\mathbb I(d_i^k\leq 3) \frac{\hat{\bm{M}}_i}{d_i^k}\frac{m_i}{d_i^k}x_i,
\end{equation}
where $x_i$ are the cell positions within the heatmap $\hat{\bm M}$, $d_i^k$ is the distance between cell $x_i$ and centroid $c_k$, $m_i = \min_{k} d_i^{k}$ is the distance from $x_i$ to its closest centroid, and $N_k$ is a normalization factor.

Unlike classical $k$-means, which updates each centroid using only the points currently assigned to it (regardless of their distance), HOME includes cells from all clusters within a radius of \SI{3}{\meter} around the centroid to allow for flexible decision boundaries. Additionally, they apply a custom weighting factor per cell based on the ratio between its distance to the closest centroid and its distance to the centroid being updated. These heuristics are reasonable when designing an end-to-end system tuned for Argoverse, but the resulting endpoints cannot be interpreted as solutions to the minFDE optimization problem under an arbitrary predictive distribution. Instead, these design choices are tightly coupled to their specific model and therefore cannot serve as a general, model-agnostic policy for other forecasting approaches. In contrast, our policy directly optimizes minFDE under the predictive distribution $p_{nT}$ by choosing endpoints that minimize the objective in Eq.~(4). It is therefore directly tied to the metric definition and only requires the ability to sample from the predictive distribution.

\section{Policy for minADE}

\label{app:ade}

The minimum average displacement error (minADE) is defined analogously to minFDE, but averages the error over all timesteps $t \in \{1, \dots, T\}$:
\begin{equation}
    \text{minADE}(\bm Y_{n}^\text{pos}, \{\hat{\bm X}_{nk}^\text{pos}\}_{k=1}^K) = \min_k \frac{1}{T}\sum_{t=1}^T\|\bm y_{nt}^\text{pos} - \hat{\bm x}_{nkt}^\text{pos}\|_2.
    \label{eq:minade}
\end{equation}

In principle, we can apply the same approach as for minFDE, except that we need to optimize full trajectories $\{\hat{\bm X}_{nk}^\text{pos}\}_{k=1}^K$, where each trajectory is a sequence $(\hat{\bm x}_{nkt}^\text{pos})_{t=1}^T$. This requires models to define a predictive distribution over complete trajectories from which we can draw an arbitrary number of samples. Many current architectures do not support this directly: they either parametrize independent per-timestep distributions \cite{knoche2025donut, lin2024eda, shi2024mtrv3, shi2024mtrpp, sun2025rmpyolo, sun2025impact, yan2025trajflow}, or they directly regress a fixed set of $K$ trajectories \cite{lan2024sept, zhang2025polaris, zhang2024demo}. For this reason, we focus on endpoint-based metrics, which only require the endpoint marginal and are therefore more generally applicable across existing forecasting models.

\section{Implementation Details}

\label{app:impl}

\PAR{Multi-horizon aggregation.}
The Waymo benchmark evaluates all metrics at \SI{3}{\second}, \SI{5}{\second}, and \SI{8}{\second} and averages the results. Accordingly, we apply the corresponding policies to the predictive endpoint distribution at these three horizons separately, and then combine the selected endpoints into a single trajectory for each mode. Intermediate timesteps do not affect these metrics and are therefore ignored.

A practical difficulty arises because confidences for (soft) mAP must be defined per trajectory for submission to the Waymo benchmark. However, our policies operate per horizon and produce separate confidences for each endpoint. To obtain per-trajectory confidences, we first sort the $K$ candidate endpoints by their policy-derived confidence in descending order for each horizon. We then construct $K$ trajectories by pairing endpoints with the same rank across horizons: trajectory $k$ consists of the $k$-th most confident endpoint at \SI{3}{\second}, \SI{5}{\second}, and \SI{8}{\second}. This rank-wise pairing ensures that the endpoints belonging to the same trajectory correspond to similar confidence levels across horizons, so that the single confidence we later assign to the trajectory remains similar to the individual confidences produced by the per-horizon policies.

We then assign a single confidence to each trajectory. Rather than averaging the three per-horizon confidences, we choose the confidence of the \SI{8}{\second} endpoint as the trajectory confidence. The longest horizon is both the most challenging and the most informative part of the forecasting task, and it typically dominates the practical usefulness of the predictions. Using the \SI{8}{\second} confidence therefore makes the trajectory-level ranking used by (soft) mAP as faithful as possible to the policy we would ideally apply at the most important horizon, while keeping the evaluation protocol simple and deterministic.

Note that this issue does not arise for minFDE, as confidences do not affect the metric. Any assignment of endpoints to trajectories yields the same minFDE.

Since intermediate timesteps are not evaluated when submitting to the benchmarks, we keep them undefined. As many current models can only produce distributions or samples for individual timesteps, full trajectories can only be produced by post-processing (\eg, interpolation).

\PAR{Policy hyperparameters.} We use $|S| = 3000$ samples for all Monte Carlo sets. To optimize minFDE, we initialize the $K=6$ endpoints by sampling from $S$ and apply the Adam optimizer \cite{kingma2015adam} for 300 steps with a learning rate of 0.2. We repeat this procedure 10 times with different random initializations and select the solution with the lowest empirical objective value.

\PAR{DONUT-NLL.} For DONUT-NLL, we mostly use the same architecture and training parameters as DONUT \cite{knoche2025donut}, but add the new loss variants. We preprocessed Waymo such that the resulting structure is similar to DONUT's preprocessing for Argoverse, but added stop signs and traffic lights (with their last observed state) as additional polygons. To decrease the scene size, we ignore stationary agents that are farther than \SI{3}{\meter} from the road, we remove all driveways and road polygons far away from the target agent, and we only keep every fifth point of each map polyline. As DONUT requires full observed history, we apply PCHIP interpolation to obtain full agent histories. 

\section{Position Distributions}

\label{app:distributions}

The Laplace distribution is parameterized by a location $\mu \in \mathbb R$ and a scale $b > 0$ as
\begin{equation}
    \text{Laplace}(x \mid \mu, b) = \frac{1}{2b}\exp\left[-\frac{|x-\mu|}{b}\right].
\end{equation}

With mean $\mu$ and standard deviation $\sigma > 0$, the Gaussian distribution is defined as
\begin{equation}
    \mathcal N(x \mid \mu, \sigma^2) = \frac{1}{\sqrt{2\pi}\sigma} \exp\left[-\frac{1}{2}\left(\frac{x-\mu}{\sigma}\right)^2\right].
\end{equation}

The generalized Gaussian uses a location $\mu \in \mathbb R$, a scale $\alpha > 0$, and a shape parameter $\beta > 0$, and is given by
\begin{equation}
    \text{GG}(x \mid \mu, \alpha, \beta) = \frac{\beta}{2\alpha\Gamma(1/\beta)}\exp\left[-\left|\frac{x-\mu}{\alpha}\right|^\beta\right],
\end{equation}
where $\Gamma(\cdot)$ denotes the Gamma function. Laplace and Gaussian distributions are special cases of the generalized Gaussian family with $\beta=1$ and $\beta=2$, respectively.

We also use discrete scale mixtures of Gaussians with a fixed number of $J=5$ components that share a single mean but differ in their scale. We define
\begin{equation}
\text{SMG}(x \mid \mu, \{\omega_j, \sigma_j\}_{j=1}^J)
= \sum_{j=1}^J \omega_j \, \mathcal N(x \mid \mu, \sigma_j^2),
\end{equation}
where $\mu \in \mathbb R$ is the shared location parameter, $\sigma_j > 0$ are component standard deviations, and $\omega_j \geq 0$ are mixture weights satisfying $\sum_{j=1}^J \omega_j = 1$.

We use each of these distribution families as a direct substitute for the 1D longitudinal and lateral Laplace distributions in DONUT. While the number of parameters needed to specify the distribution differs between families, this only changes the dimensionality of the model’s output head and can be implemented by adjusting the size of the final linear layer at each decoding step.

\section{Full Ablations}

\label{app:full_abl}

In \cref{tab:distr_ablations}, we show the full ablations with all losses and distribution families. Overall, applying our TraDiE policies improves performance compared to naive evaluation, except for WTA, where minFDE degrades. This is consistent with the fact that WTA explicitly optimizes for the endpoint error rather than a properly calibrated predictive distribution.

We further observe that optimizing for the quality of the predictive distribution leads to different preferred design choices than optimizing directly for benchmark metrics. With our policies applied, Step-NLL combined with the generalized Gaussian consistently outperforms all other approaches, whereas naive evaluation would favor different losses and position distributions.

\begin{table*}[t]
    \centering
    \smalltext
    \renewcommand{\tabcolsep}{5pt}
    \caption{\textbf{Ablation on position distributions.} Applying our policies for distance- or window-based metrics generally outperforms the naive predictions (\textbf{bold}). Notably, naive evaluation suggests a different best training objective and position distribution than the policy-optimized evaluation (\bestnaive{\phantom{X}} vs.\ \bestpolicy{\phantom{X}}). Numbers are shown in \hide{grey} when they are obtained using policies that are not appropriate for the corresponding metric. Evaluation on Waymo~\cite{ettinger2021waymo} \textit{val}.}
    \label{tab:distr_ablations}
    \begin{tabular}{lllrrrrrr} \toprule
        Opt. & Loss      & Pos.-Distr.    & Soft mAP \up & mAP \up & Miss rate \down & minFDE \down  \\
        \midrule
        \multirow{9}{*}{\rotatebox{90}{naive}}
             & WTA       & Laplace        & 0.3427 & 0.2995 & 0.1310 & \best{1.0520} \\
             & WTA       & Gen.\ Gaussian & 0.3476 & 0.3071 & 0.1291 & \best{1.0594} \\
             & WTA       & Scale Mixture  & 0.3579 & 0.3154 & \bestnaive{0.1268} & \bestnaive{\best{1.0428}} \\
             & Traj-NLL & Laplace        & 0.3401 & 0.2851 & 0.1286 & 1.0874 \\
             & Traj-NLL & Gen. Gaussian  & 0.3481 & 0.2923 & 0.1297 & 1.1120 \\
             & Traj-NLL & Scale Mixture  & 0.3639 & 0.3084 & 0.1276 & 1.0862 \\
             & Step-NLL  & Laplace        & 0.3595 & 0.3163 & 0.1444 & 1.2236 \\
             & Step-NLL  & Gen.\ Gaussian & 0.3759 & 0.3439 & 0.1561 & 1.3127 \\
             & Step-NLL  & Scale Mixture  & \bestnaive{0.3837} & \bestnaive{0.3533} & 0.1492 & 1.2722 \\
        \midrule
        \multirow{9}{*}{\rotatebox{90}{window policy}}
             & WTA       & Laplace        & \best{0.4764} & \best{0.4736} & \best{0.1044} & \hide{1.3466} \\
             & WTA       & Gen.\ Gaussian & \best{0.4764} & \best{0.4733} & \best{0.1059} & \hide{1.3619} \\
             & WTA       & Scale Mixture  & \best{0.4827} & \best{0.4798} & \best{0.1014} & \hide{1.3640} \\
             & Traj-NLL & Laplace        & \best{0.4921} & \best{0.4892} & \best{0.0946} & \hide{1.3317} \\
             & Traj-NLL & Gen. Gaussian  & \best{0.4825} & \best{0.4794} & \best{0.0975} & \hide{1.3669} \\
             & Traj-NLL & Scale Mixture  & \best{0.4963} & \best{0.4935} & \best{0.0947} & \hide{1.3418} \\
             & Step-NLL  & Laplace        & \best{0.5082} & \best{0.5053} & \best{0.0902} & \hide{1.3100} \\
             & Step-NLL  & Gen.\ Gaussian & \bestpolicy{\best{0.5101}} & \bestpolicy{\best{0.5070}} & \bestpolicy{\best{0.0876}} & \hide{1.2980} \\
             & Step-NLL  & Scale Mixture  & \best{0.5053} & \best{0.5024} & \best{0.0909} & \hide{1.3098} \\
        \midrule
        \multirow{9}{*}{\rotatebox{90}{distance policy}}
             & WTA       & Laplace        & \hide{0.1593} & \hide{0.1103} & \hide{0.1325} & 1.0824 \\
             & WTA       & Gen.\ Gaussian & \hide{0.1584} & \hide{0.1102} & \hide{0.1306} & 1.0884 \\
             & WTA       & Scale Mixture  & \hide{0.1626} & \hide{0.1115} & \hide{0.1295} & 1.0808 \\
             & Traj-NLL & Laplace        & \hide{0.1626} & \hide{0.1142} & \hide{0.1202} & \best{1.0374} \\
             & Traj-NLL & Gen. Gaussian  & \hide{0.1609} & \hide{0.1129} & \hide{0.1221} & \best{1.0477} \\
             & Traj-NLL & Scale Mixture  & \hide{0.1639} & \hide{0.1136} & \hide{0.1206} & \best{1.0344} \\
             & Step-NLL  & Laplace        & \hide{0.1708} & \hide{0.1148} & \hide{0.1186} & \best{1.0200} \\
             & Step-NLL  & Gen.\ Gaussian & \hide{0.1689} & \hide{0.1146} & \hide{0.1169} & \bestpolicy{\best{1.0103}} \\
             & Step-NLL  & Scale Mixture  & \hide{0.1707} & \hide{0.1145} & \hide{0.1195} & \best{1.0218} \\
        \bottomrule
    \end{tabular}
\end{table*}

\section{Limitations}

\subsubsection{Continuity of trajectories.}
A clear limitation of our minFDE and mAP policies is that they generally do not produce temporally continuous trajectories. Downstream applications that require $K=6$ continuous trajectories that are themselves optimal for minFDE or mAP cannot use these policies directly. However, applicability to downstream control or planning is not the purpose of these particular policies. Our goal is to evaluate the quality of the predictive distribution under the metrics used by the benchmarks. Given a calibrated predictive distribution, any downstream task can define its own policy tailored to that specific use case.

\subsubsection{Runtime.}
With default parameters, optimizing (soft) mAP with our policy takes approximately \SI{0.03}{\second} per scene, while minFDE is considerably slower at around \SI{0.26}{\second} per scene. This additional cost is incurred only at evaluation time.

Note that we did not aim for efficiency in this work. Our policies are designed for evaluation, not as AV planning-stack components, and they serve as a proof of concept for tailoring policies to a specific downstream task.

\section{Visualizations}

We visualize endpoint distributions together with naive endpoints, endpoints optimized for minFDE, and endpoints optimized for soft mAP in \cref{fig:scene_0,fig:scene_1,fig:scene_2,fig:scene_3,fig:scene_4,fig:scene_5,fig:scene_6,fig:scene_7,fig:scene_8,fig:scene_9} for DONUT and DONUT-NLL. In general, endpoints optimized for soft mAP show greater coverage than endpoints obtained under the minFDE policy. In some cases, they even lie on individual outliers when the windows in the main mode of the distribution are already covered by higher-confidence endpoints (\cref{fig:scene_0}). Such outlier endpoints receive low confidences from our policy and therefore have little effect on the (soft) mAP score.

In \cref{fig:scene_9}, we can observe a limitation of winner-takes-all training. Because the straight and right-turn futures have much higher likelihood than the left-turn future, the model places all mixture components along these futures to optimize the WTA objective. At the same time, it is aware of the possibility of turning left and therefore inflates the uncertainty of the modes to account for this alternative. This results in very wide modes that do not reflect the true multimodal structure of the future and thus deviate from a well-calibrated predictive distribution.

\foreach \x in {0,...,9} {
    \newcommand{\heatmap}{%
        {\color{RWTHGelb}h}%
        {\color{RWTHGelb!50!RWTHOrange}e}%
        {\color{RWTHOrange}a}%
        {\color{RWTHOrange!50!RWTHRot}t}%
        {\color{RWTHRot}m}%
        {\color{RWTHRot!50!RWTHBordeaux}a}%
        {\color{RWTHBordeaux}p}%
    }
    \begin{figure}
        \centering
        \includegraphics[width=0.8\textwidth]{fig/im/scene_\x-wta-laplace.png}
        \includegraphics[width=0.8\textwidth]{fig/im/scene_\x-traj-nll-laplace.png}
        \includegraphics[width=0.8\textwidth]{fig/im/scene_\x-step-nll-laplace.png}
        \includegraphics[width=0.8\textwidth]{fig/im/scene_\x-step-nll-generalized-gaussian.png}
        \includegraphics[width=0.8\textwidth]{fig/im/scene_\x-step-nll-scale-mixture.png}
        \caption{\textbf{Visualization of a random scene.} The \heatmap{} shows the endpoint distribution for the target agent {\colorbox{RWTHGruen}{\phantom{XX}}}. Naive endpoints and endpoints under our policy are marked with \texttimes. \textbf{Pos-NLL Generalized Gaussian} is the best performing model after applying our policies.}
        \label{fig:scene_\x}
    \end{figure}
}

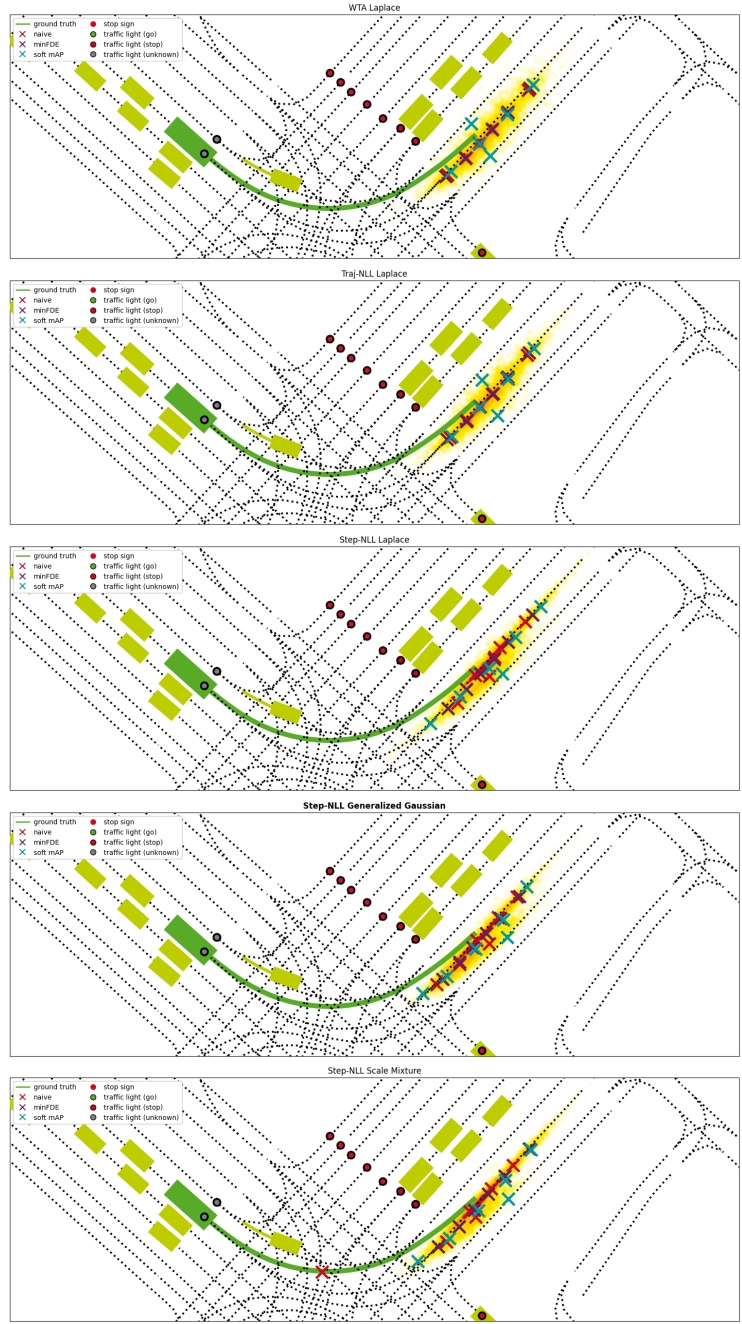
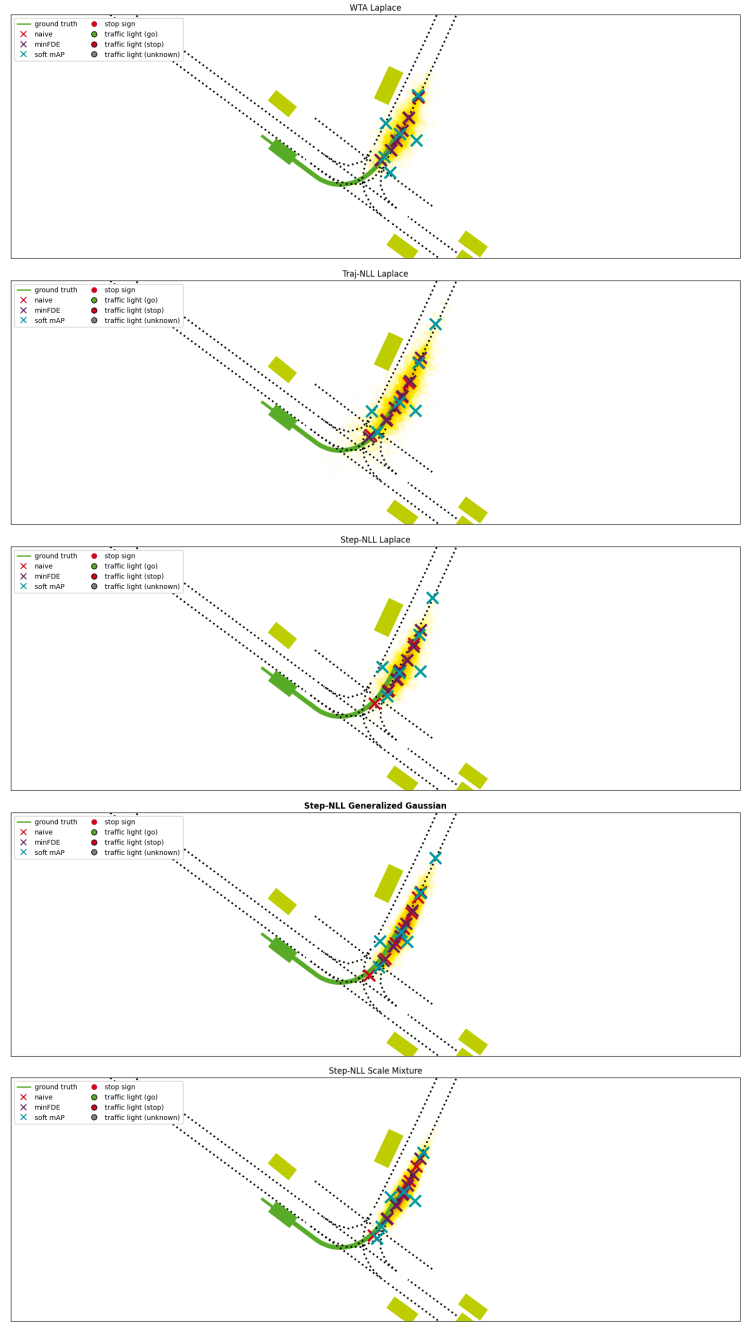
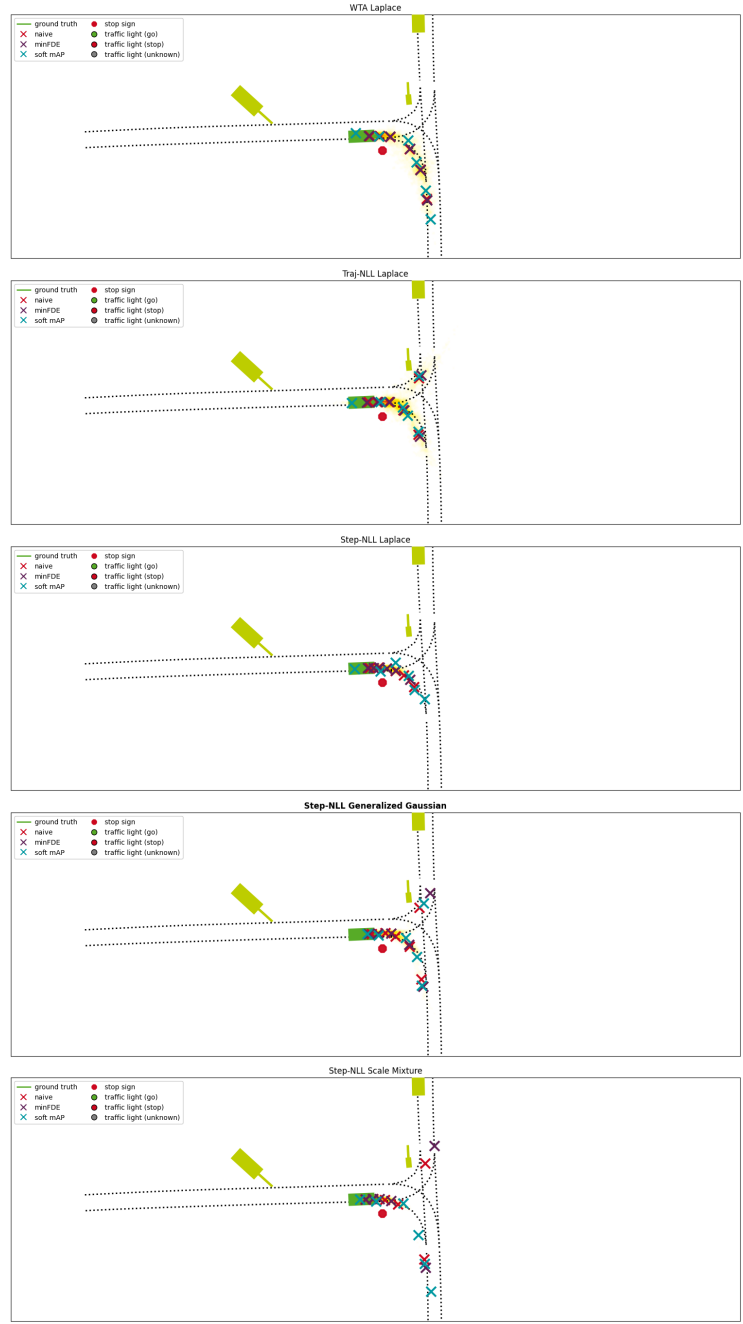
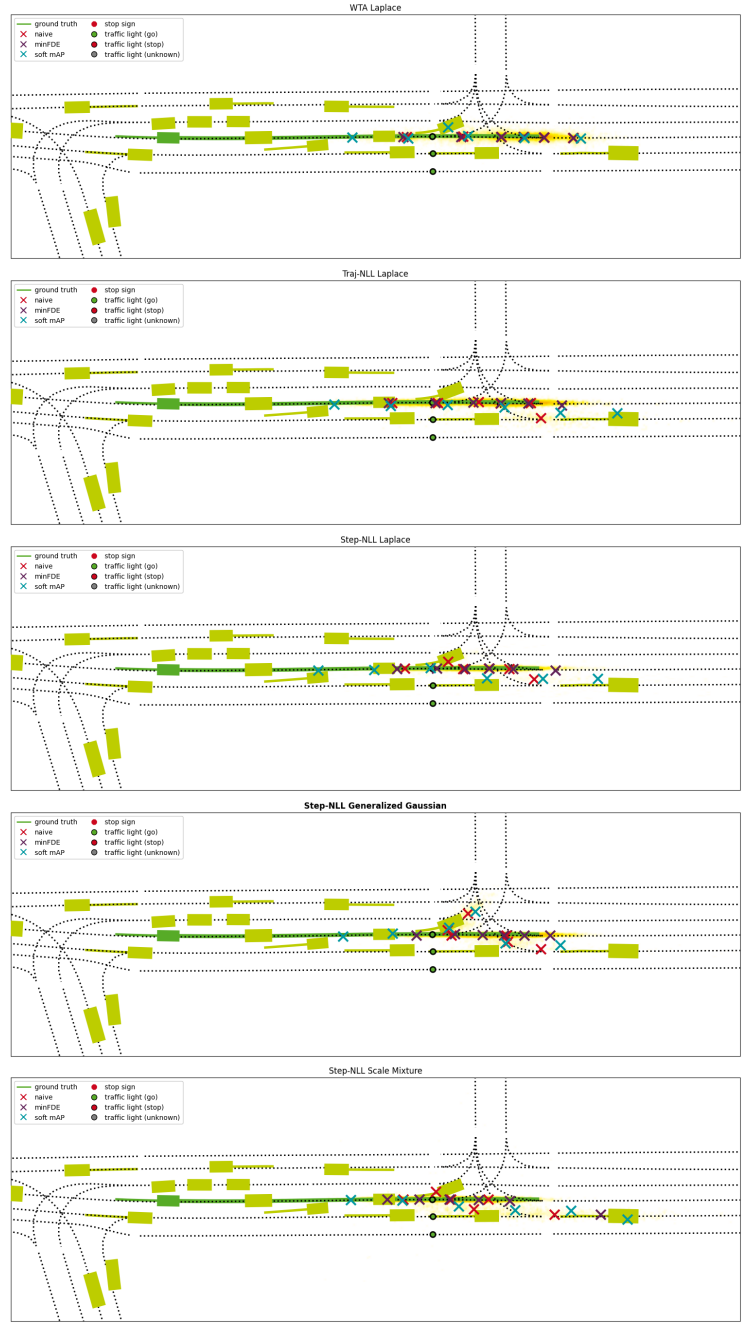
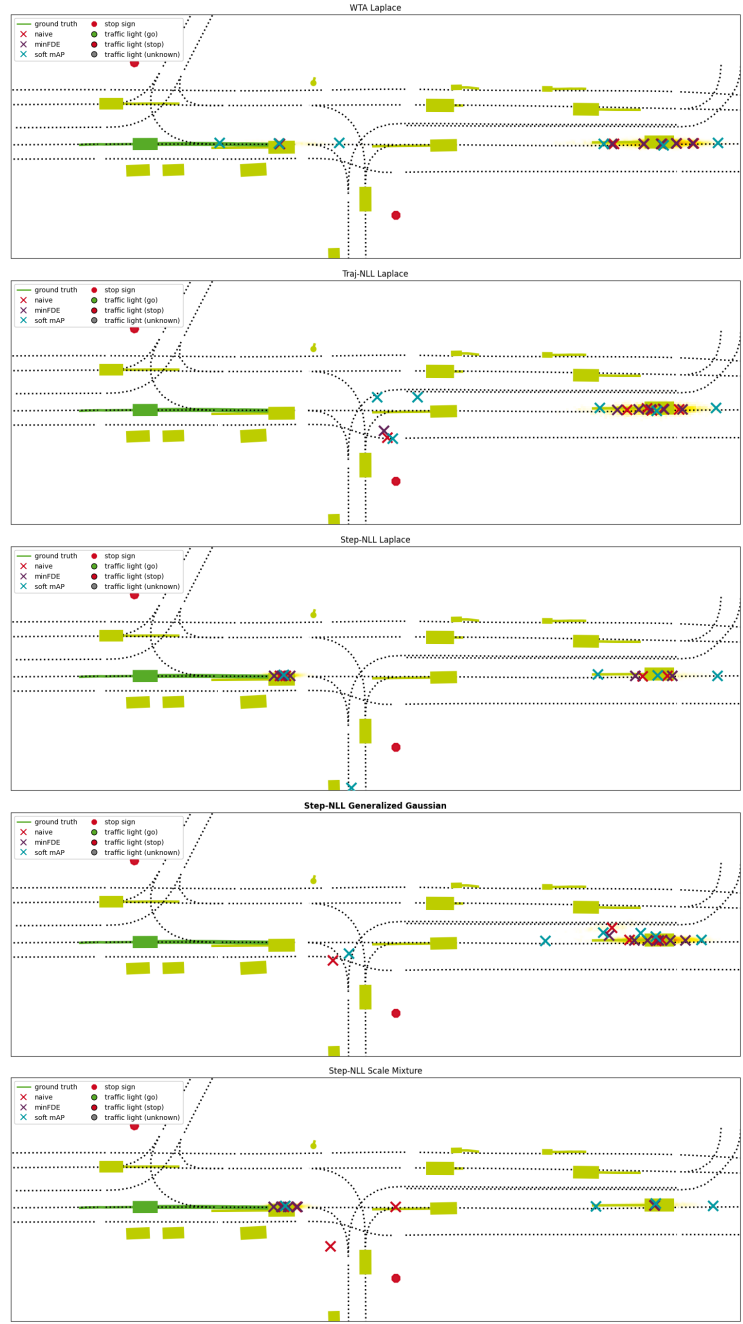
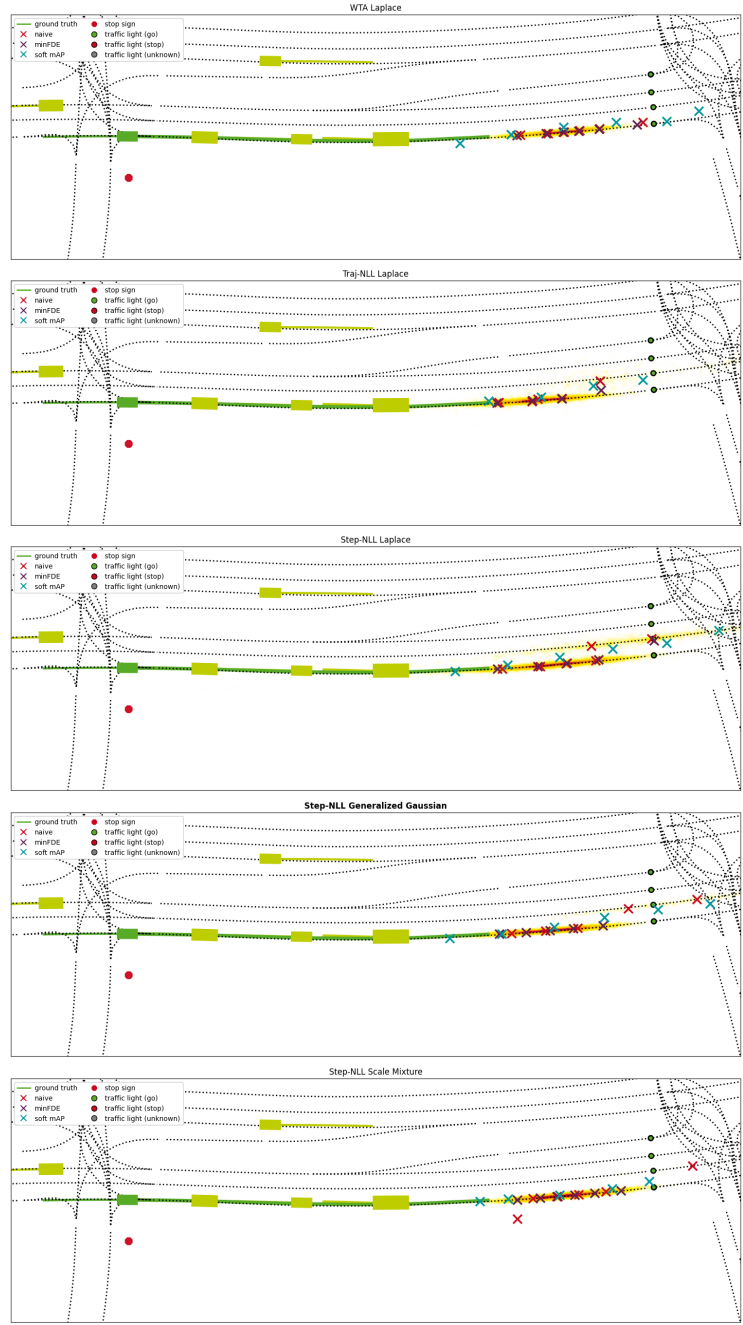
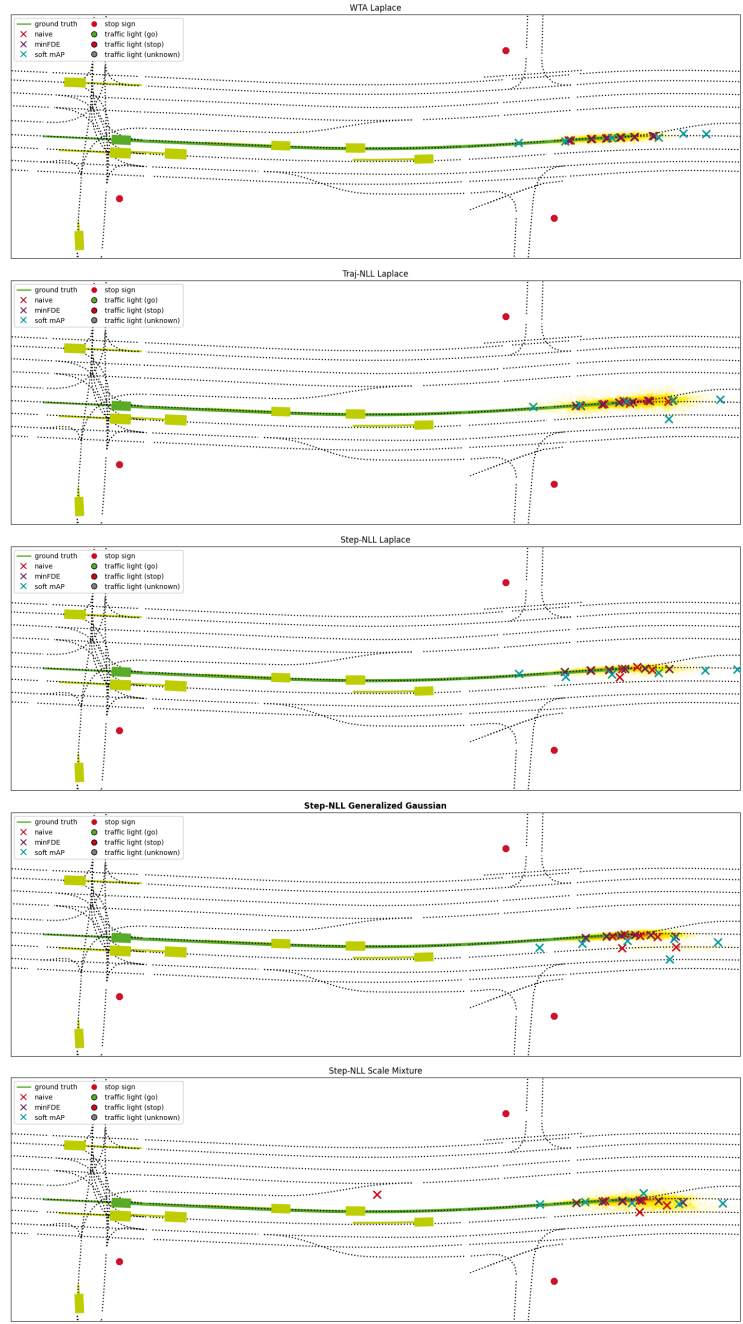
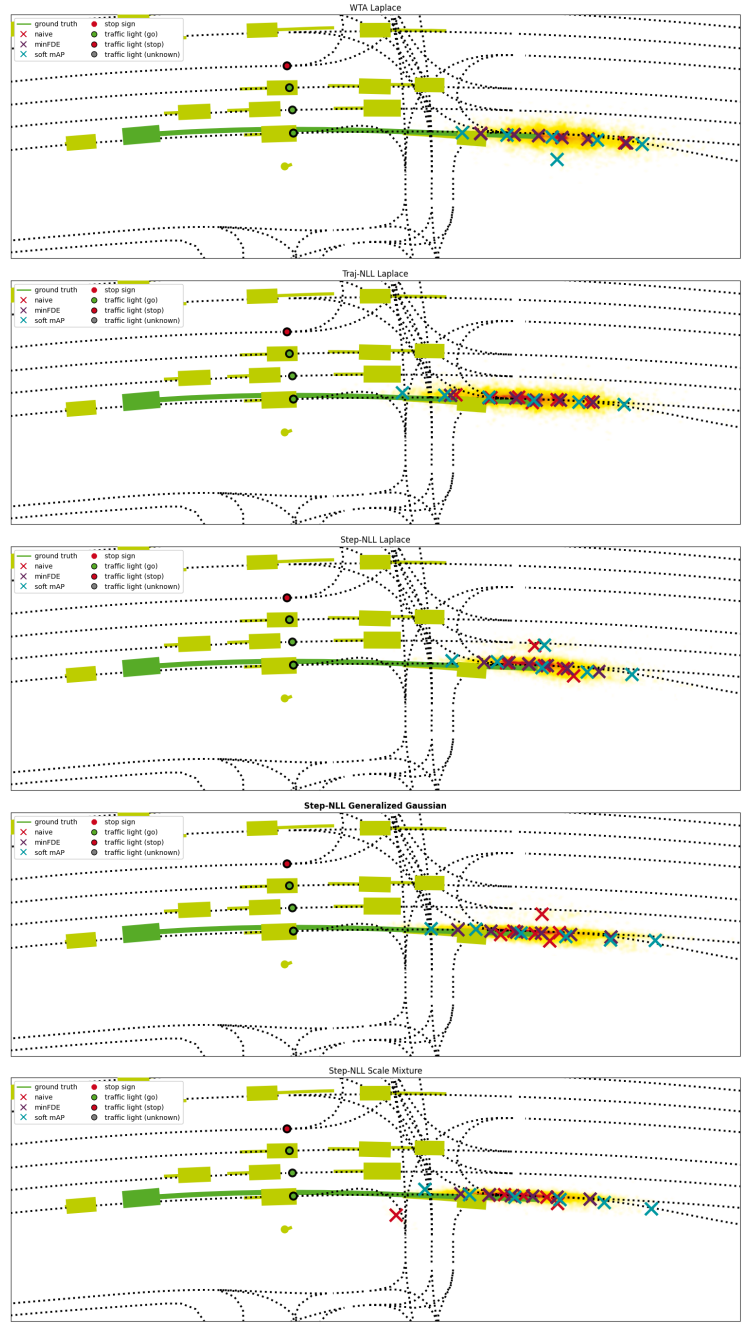
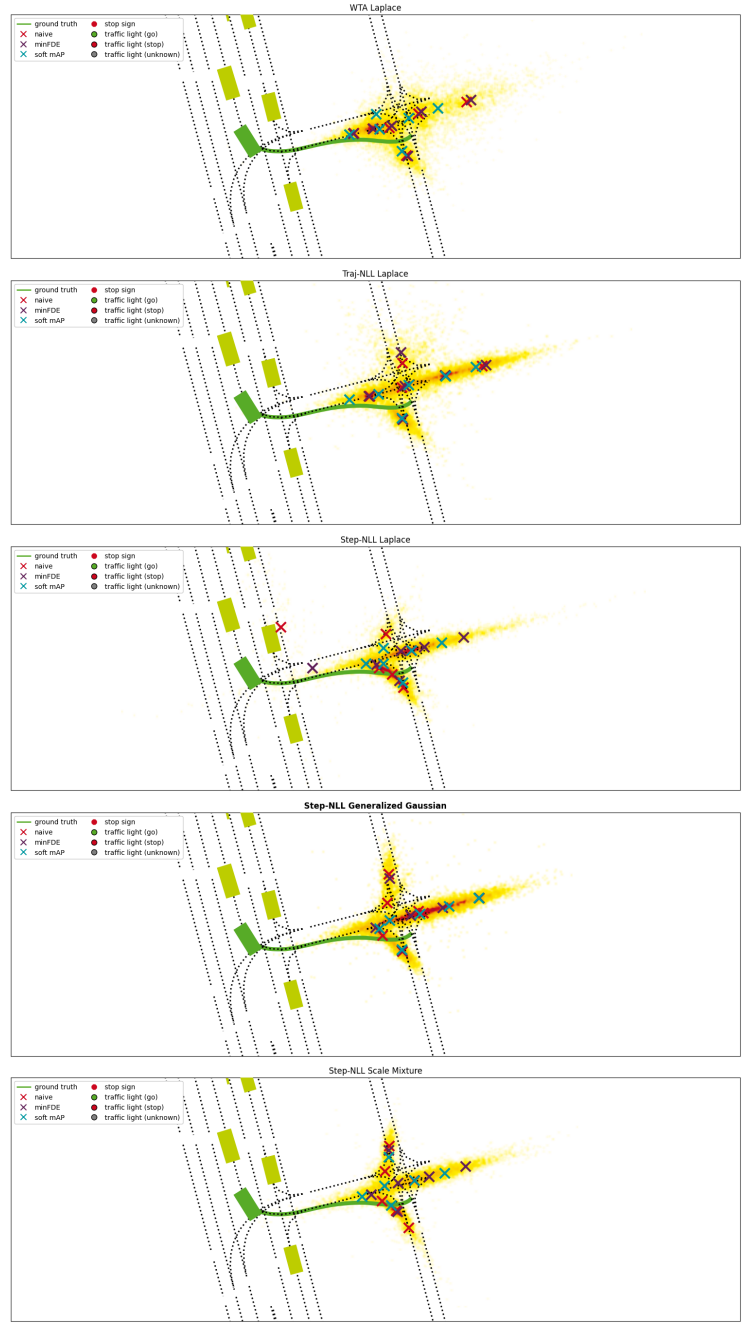
\end{document}